\documentclass[sigconf]{acmart}
\AtBeginDocument{%
  }

\copyrightyear{2026}
\acmYear{2026}
\setcopyright{cc}
\setcctype{by}
\acmConference[KDD '26]{Proceedings of the 32nd ACM SIGKDD Conference on Knowledge Discovery and Data Mining V.2}{August 09--13, 2026}{Jeju Island, Republic of Korea}
\acmBooktitle{Proceedings of the 32nd ACM SIGKDD Conference on Knowledge Discovery and Data Mining V.2 (KDD '26), August 09--13, 2026, Jeju Island, Republic of Korea}
\acmDOI{10.1145/3770855.3817733}
\acmISBN{979-8-4007-2259-2/2026/08}

\usepackage{graphicx}
\usepackage{textcomp}
\usepackage{xcolor}
\usepackage{mwe}
\usepackage{algorithm}
\usepackage{algpseudocode} 
\usepackage{dsfont}
\usepackage{subfigure}
\usepackage{newfloat}
\usepackage{listings}
\usepackage{bibentry}
\usepackage{comment}
\usepackage{amsmath,amsfonts,amsthm,bm}
\usepackage{commath}
\usepackage{multirow}
\usepackage{xcolor,colortbl}
\usepackage{ulem}

\newtheorem{pblm}{Problem}

\newcommand{\Lhse}{InfluenceCB}

\newcommand{\E}{\mathcal{E}}
\newcommand{\A}{\mathcal{A}}
\newcommand{\wb}{\mathbf{w}}

\usepackage{thmtools,thm-restate}

\theoremstyle{plain}
\newtheorem{theorem}{Theorem}[section]

\newtheorem{lemma}[theorem]{Lemma}

\theoremstyle{definition}

\theoremstyle{remark}
\newtheorem{remark}[theorem]{Remark}
\usepackage{enumitem}

\title{Learning Peer Influence Probabilities\\ with Linear Contextual Bandits}
\author{Ahmed Sayeed Faruk}
\affiliation{%
  \institution{University of Illinois Chicago}
  \city{Chicago}
  \country{USA}
}
\email{afaruk2@uic.edu}

\author{Mohammad Shahverdikondori}
\affiliation{%
  \institution{École Polytechnique Fédérale de}
  \country{Lausanne, Switzerland}
}
\email{mohammad.shahverdikondori@epfl.ch}

\author{Elena Zheleva}
\affiliation{%
  \institution{University of Illinois Chicago}
  \city{Chicago}
  \country{USA}
}
\email{ezheleva@uic.edu}

\renewcommand{\shortauthors}{Faruk, Shahverdikondori, and Zheleva}

\begin{abstract}
In networked environments, it is common for users to share recommendations about content, products, services, and possible courses of action. Whether these recommendations are accepted and acted upon is highly context-dependent, influenced by the characteristics of the sender and recipient, the nature of their relationship, the attributes of the recommended item, and the communication context. Consequently, probabilities of peer influence exhibit substantial heterogeneity across individuals and settings. Accurate estimation of these probabilities is key to understanding information diffusion processes and to improving the effectiveness of viral marketing strategies. However, learning these probabilities from data is challenging; static data may capture correlations between peer recommendations and peer actions but fails to reveal influence relationships. Online learning algorithms can learn these probabilities from interventions but either waste resources by learning from random exploration or optimize for rewards, thus favoring exploration of the space with higher influence probabilities. In this work, we study learning peer influence probabilities under a contextual linear bandit framework. We show that a fundamental trade-off can arise between regret minimization and estimation error, characterize all achievable rate pairs, and propose an uncertainty-guided exploration algorithm that, by tuning a parameter, attains any pair within this trade-off. Our experiments on semi-synthetic network datasets show the advantages of our method over static methods and contextual bandits that ignore this trade-off.
\end{abstract}


\begin{document}
\maketitle
\newcommand\kddavailabilityurl{https://doi.org/10.5281/zenodo.20500085}
\ifdefempty{\kddavailabilityurl}{}{
\begingroup\small\noindent\raggedright\textbf{Resource Availability:}\\
The source code of this paper has been made publicly available at
\url{\kddavailabilityurl}.
\endgroup
}
\section{Introduction}
Influence occurs when the action of one user affects the actions of other users in a social network, spreading information, behaviors, and attitudes. Influence is an important driver of user behavior in many digital platforms, including social media, e-commerce, and content platforms~\cite{hollebeek-ir-2023, kuang-mis-2019, xu-tm-2022}. 
Influence probability, the extent to which one user can influence another, varies with users’ network positions, connections, and individual traits. For example, in a social network, sharing a recommendation can result in adoption by some friends and in lack of interest by others. Predicting and leveraging these heterogeneous influence probabilities is very important for understanding information diffusion processes and for improving the effectiveness of viral marketing strategies. 

Estimating influence probabilities from correlations between user actions is challenging because correlations can be attributed to other potential sources, such as homophily, confounding factors, and mere coincidence~\cite{anagnostopoulos-kdd-2008, shalizi-smr-2011}. 
Homophily refers to the tendency of similar individuals to form connections and exhibit similar behaviors~\cite{newman-pr-2003}. Confounding factors, on the other hand, can simultaneously affect the formation of friendships and subsequent actions~\cite{shalizi-smr-2011,marlow-ht-2006}. 
Many influence-learning methods~\cite{goyal-wsdm-2010, kutzkov-kdd-2013,saito-kes-2008, subbian-wsdm-2016, zhang-aaai-2017, zhang-www-2019, chakraborty-wsdm-2023} do not disentangle these sources of correlations from true influence. 


Contextual multi-armed bandits (CMABs)~\cite{chen-icml-2013} 
and online learning~\cite{shalev-2012} 
algorithms have the potential to learn influence probabilities from interventions rather than pure correlations but their objectives may not always offer an optimal solution. Online learning methods can optimize for influence probability estimation error by exploring at random, but this approach is resource-inefficient and lacks regret guarantees. CMABs offer a principled framework for sequential, context-aware interventions but they aim to minimize cumulative regret, which may limit exploration to the regions with high influence probabilities, this resulting in high influence probability estimation error. 
Existing research on multi-armed bandits in networked settings has considered leveraging heterogeneous influence probabilities, assuming that they are known, for tasks such as predicting recommendations~\cite{faruk-arxiv-2023} and influence maximization~\cite{iacob-sdm-2022,wilder-aaai-2018}. Others learn the influence probabilities in the context of influence maximization, thus focusing on learning and leveraging high influence probabilities~\cite{vaswani-icml-2017,wen-neurips-2017}. More recently, there are studies in the setting when treatments are simultaneously assigned to all nodes and rewards depend on neighbors’ actions~\cite{mab-interference2-jia2024multi, mab-interference1-agarwal2024mutli, mab-interference3-jamshidi2025graph, mab-interference4-xu2024linear, mab-trade-off-zhang2024online}. However, none of these studies address the task of learning peer influence probabilities as an independent problem. 


In this paper, we study the problem of learning peer influence probabilities from $k$ edge interventions per round, where interventions may be selected either from the whole network or from the neighbors of specific nodes. An intervention refers to an online platform showing the action of one user (e.g., social media post or product recommendation) to another user to see whether the recipient would take a subsequent action on it (e.g., share the post or buy the product). 
The peer influence probability corresponds to the probability that the peer will take a subsequent action. 
We focus on actions that are visible to the peer (e.g., the post/recommendation is public) and use features that do not have privacy concerns (e.g., from publicly available information). 
We cast this problem as a linear contextual bandit task and prove the existence of a fundamental trade-off between cumulative regret and influence probability estimation error.  
Analyzing this trade-off in our setting introduces additional challenges compared to standard bandit problems. These challenges arise from the combinatorial structure of the action space: the number of edges in the network, and hence the number of influence probabilities to be estimated, is large, and at each round the agent observes activation outcomes for multiple edges simultaneously.
We characterize all achievable rate pairs for this trade-off and show that no algorithm can achieve the optimal rates for both objectives simultaneously. Modeling influence with a linear bandit allows us to give trade-off guarantees and is consistent with existing diffusion literature that treats social influence as an additive aggregation of contextual signals~\cite{kempe-kdd-2003, aral-pnas-2009, he-neurips-2016}. 

Building on this theory, we propose \textbf{Influence Contextual Bandit} ($\Lhse$), a framework that alternates between uncertainty-guided exploration and reward-oriented exploitation (Figure 1). $\Lhse$ uses an objective-driven uncertainty threshold to decide when to explore uncertain edges versus exploit high-probability edges and employs a parameter $\beta$ to navigate the trade-off frontier. $\Lhse$ differs from standard contextual bandits because in addition to optimizing rewards, it has a second objective, minimizing global estimation error and using $\beta$ to enforce systematic exploration of low-influence edges. Our framework can incorporate various features, including node, edge, and network attributes. 

\textbf{Key idea and highlights.}
To summarize, this paper makes the following contributions: 
\begin{itemize}[topsep=2pt]
  \item We formulate the problem of learning heterogeneous peer influence probabilities in networks as a bi-objective contextual multi-armed bandit, aiming to simultaneously minimize cumulative regret and estimation error. 
  \item We theoretically establish a fundamental trade-off between regret minimization and estimation error and characterize the achievable rate pairs.
  \item We propose $\Lhse$, a CMAB framework that incorporates uncertainty-guided exploration to accelerate influence probability learning while minimizing resource waste.
  \item We provide theoretical guarantees for linear settings and show that $\Lhse$ can achieve any point on the trade-off curve by tuning a single parameter.
  \item We evaluate $\Lhse$ on semi-synthetic network datasets and show its effectiveness over methods based on observational data or conventional contextual bandits.
\end{itemize}
\section{Related Work}







Link weight prediction focuses on estimating the numerical value (e.g., strength, capacity) associated with an edge in a network~\cite{fu-ieeetkdd-2018}. Previous studies have explored influence probability estimation either offline from historical cascades using propagation models or likelihood-based inference~\cite{goyal-wsdm-2010,kutzkov-kdd-2013,saito-kes-2008,chakraborty-wsdm-2023}, or have focused on structural patterns~\cite{zhang-aaai-2017} and influence tracking or maximization without learning edge-level probabilities~\cite{subbian-wsdm-2016,zhang-www-2019}. 
In contrast, our work learns edge-level influence probabilities online through interventions, explicitly modeling the exploration–exploitation trade-off. Although we do not model cascades or multi-hop diffusion, the learned probabilities can potentially be used in diffusion simulations and influence maximization algorithms.



Learning influence probabilities with bandits has been studied for problems such as probabilistic maximum coverage and social influence maximization in viral marketing~\cite{vaswani-icml-2017,wen-neurips-2017}. However, they are interested in learning and leveraging high influence probabilities, rather than the full range of probabilities. 
A related direction is multi-armed bandits with network interference~\cite{mab-interference2-jia2024multi,mab-interference1-agarwal2024mutli,mab-interference3-jamshidi2025graph,mab-interference4-xu2024linear,mab-trade-off-zhang2024online}, where treatments are assigned simultaneously across all nodes and rewards depend on neighbors’ actions.  Fundamental trade-offs between cumulative regret and learning objectives have been studied in this setting, including estimation accuracy in bandits with interference~\cite{mab-trade-off-zhang2024online}, best arm identification~\cite{bai-trade1-zhongachieving,bai-trade2-degenne2019bridging}, and graph learning in causal bandits~\cite{graph-trade-shahverdikondori2025graph}. In contrast, our framework considers the trade-off between regret and learning peer influence probabilities by selecting a small subset of edges per round and observing only their local influence.

We adopt CombLinUCB-style
guarantees~\cite{comblinucb-wen2015efficient} as our regret subroutine
because they suffice for our bi-objective analysis, while recent
advances in combinatorial CMABs have produced tighter regret bounds
under additional structural
assumptions~\cite{takemura-aaai-2021}. $C^2$MAB-T~\cite{liu-icml-2023} operates under an
Action$\rightarrow$Triggering$\rightarrow$Observation feedback model
that is structurally incompatible with our direct intervention setting
(Action$\rightarrow$Observation).

While extensive research exists on top-$k$ recommendation problems (e.g., ~\cite{jamali-recsys-2009, song-icdm-2015, yang-recsys-2012}), including with CMABs~\cite{aramayo-ms-2023}, limited attention has been given to identifying top-$k$ edges in social networks. A recommendation model for socialized e-commerce focuses on recommending top-$k$ relevant products to a user for sharing with all neighbors~\cite{gao-ieeetkde-2022} but no work focuses on recommending top-$k$ neighbors for sharing a fixed product.
\begin{figure}[t]
\centerline{\includegraphics[width=\columnwidth]{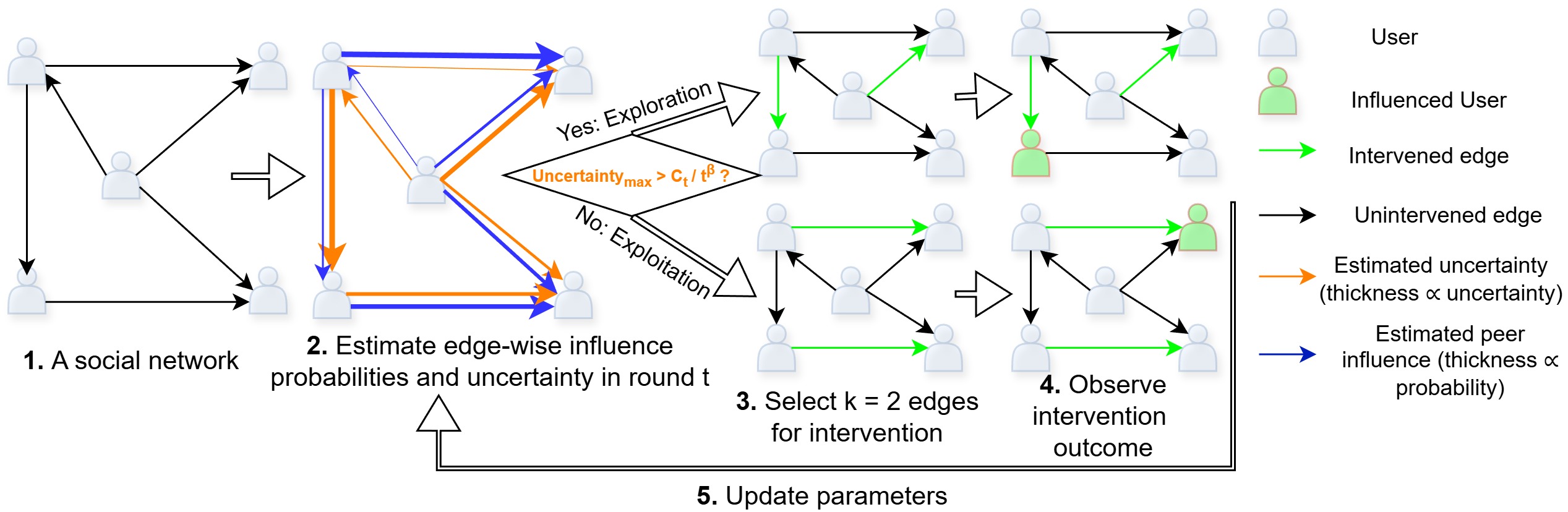}}
\caption{The workflow of the $\Lhse$ framework.}
\label{workflow}
\vspace{-1.5em}
\end{figure}

\section{Problem Description}
\label{CMAB-setup}
We consider a setting in which an online social platform deploys an influence probability learning agent to decide which peer-to-peer exposures to trigger, such as showing a post, recommendation, or referral from one user to another. Each round proceeds as follows: the agent observes a pool of candidate directed edges (e.g., between a source user who is sharing a recommendation and all their neighbors who are potential recipients) and selects a subset of $k$ edges for intervention, (e.g., $k$ recipient users who are shown the shared recommendation by the platform). After the exposure, the agent observes whether each recipient adopts the action (e.g., reshares the post), which we interpret as a peer influence event. 
This feedback is then used to update the agent’s model before the next round. From an online learning perspective, each exposure is an action with uncertain reward, and the agent must balance two competing objectives: maximizing short-term adoption while accurately learning heterogeneous peer influence probabilities across the network. 

We represent the data as a directed attributed network \( G = (V, E) \), where \( V = \{v_1, v_2, \ldots, v_n\} \) is the set of \( n \) nodes, and \( E \) is the set of edges. We denote by \( e_{ij} \), the edge between nodes \( v_i \) and \( v_j \). The neighborhood of a node \( v_i \) is defined as \( \mathcal{N}_i = \{v_j \in V \mid e_{ij} \in E\} \). The attribute vector \( e_{ij}.X\) is referred to as the context vector for edge \( e_{ij} \in E \). It includes attributes of nodes $v_i$ and $v_j$ and other relevant information, such as the recommended item attributes and network structural features (e.g., number of common neighbors). 
Each edge \( e_{ij} \in E \) is associated with a peer influence probability \( e_{ij}.p \in [0, 1] \), which represents the probability of activating node \( v_j \) (recipient node) due to peer influence from node \( v_i \) (source node). The peer influence probability can be asymmetric (\( e_{ij}.p \neq e_{ji}.p \)) and heterogeneous across edges (\( e_{ij}.p \neq e_{il}.p, e_{il}.p \neq e_{jl}.p \)). In our problem setup, these probabilities are unknown.

We consider a stochastic $M$-armed contextual linear bandit setting over $T$ rounds. The set of arms is denoted by $\mathcal{A} = \{ X_1, X_2, \ldots, \\ X_M \} \subseteq \mathbb{R}^d$, where each $X_m$ corresponds to the context vector for an edge $e_{i,j}$ and $M$ corresponds to the number of edges. At each round $t \in \{1, \dots, T\}$, the agent observes a pool of available actions $\mathcal{A}_t \subseteq \mathcal{A}$ and selects $k$ actions $\{X_{t,1}, \dots, X_{t,k}\}$ where $k \le |A_t|$ and $X_{t,i}$ denotes the $i$-th action chosen at round $t$. The action pool $\mathcal{A}_t$ may include edges from one node’s neighborhood or from multiple nodes' neighborhoods, and its size may vary over different rounds. After receiving the binary rewards $\{r_{t,1}, \dots, r_{t,k}\}$, the agent updates its selection policy. Each reward corresponds to a successful activation event: a reward of $1$ indicates that a recipient node $v_j$ was activated due to influence from a source node $v_i$ across edge $e_{i,j}$, and $0$ otherwise. 

We assume a linear reward model with an unknown parameter $\theta^* \in \mathbb{R}^d$, such that
$$
\mathbb{E}[r_{t,m}] = X_{t,m}^{\top}\theta^*.
$$

The estimated peer influence probability thus corresponds to the probability of observing reward $1$. The set of top-$k$ edges with the highest reward in round $t$ is defined as 
$$
\mathcal{E}_t^* \leftarrow \underset{X_t \in \A_t}{\arg\text{top}_k}\; X_t^{\top}\theta^*.
$$
Hence, any problem instance is characterized by $(\A, \theta^*)$.  
We further assume bounded norms for actions and parameters: there exist constants $c, c' > 0$ such that $\|X\|_2 \le c$ for all $X \in \A$ and $\|\theta^*\|_2 \le c'$. We also assume that the set of actions spans $\mathbb{R}^d$. These are standard assumptions in the linear bandit literature~\cite{bandit-book-lattimore2020bandit}. The set of all problem instances satisfying these conditions with $d$-dimensional actions and binary rewards is denoted by $\E$. We can now define our two objectives of interest.

\textbf{Regret.} The expected cumulative regret of a policy $\pi$ interacting with an environment characterized by $(\A, \theta^*)$ over $T$ rounds is defined as
$$
\mathrm{Regret}_T(\pi, \A, \theta^*) 
\;=\; 
\sum_{t=1}^{T} 
\left(
\sum_{X \in \mathcal{E}^{*}_t} X^{\top}\theta^*
\;-\;
\sum_{i=1}^k X_{t,i}^{\top}\theta^*
\right),
$$
where $\mathcal{E}^{*}_t$ denotes the set of top-$k$ optimal actions at round $t$. The goal in standard contextual bandits is to minimize $\mathrm{Regret}_T(\pi, \A, \theta^*)$.  
Since $\theta^*$ is unknown during the learning process, different bandit algorithms employ various strategies to minimize regret by selecting high-reward actions $X_{t,i}$.  
However, focusing solely on regret minimization does not necessarily lead to an accurate estimation of peer influence probabilities.  
We therefore introduce a second objective that explicitly measures estimation accuracy.  

\textbf{RMSE.} The expected root mean squared error (RMSE) of the estimated influence probabilities over all edges serves as our second objective. For a policy $\pi$, its value after $T$ rounds is defined as
$$
\mathrm{RMSE}_T(\pi, \A, \theta^*) 
\;=\; 
\mathbb{E} \left[ \sqrt{\frac{1}{|\A|} 
\sum_{X \in \A} 
\big(X^{\top} (\hat{\theta}_T - \theta^*)\big)^2} \right],
$$
where $\hat{\theta}_T$ denotes the parameter estimate obtained after $T$ rounds.

\paragraph{Why a global RMSE?}
Defining RMSE over all edges--including those rarely activated--reflects
the objective of learning the influence parameters themselves,
not merely the quality of actions taken during learning. Accurate global
estimates enable downstream uses that pure regret minimization does not:
simulating diffusion under counterfactual seedings, evaluating new
policies offline without re-exploration, auditing influence patterns
across user subgroups, and supporting influence-maximization or
targeting tasks. Restricting RMSE to high-reward edges would collapse
the problem to standard regret minimization and remove the very
tension we set out to characterize.

The two objectives—minimizing cumulative regret and minimizing estimation error—are inherently in conflict. Minimizing regret requires focusing on high-reward edges, biasing exploration toward frequently activated neighbors and leaving large regions of the influence-probability space underexplored. Conversely, reducing estimation error necessitates exploring uncertain or low-reward edges, which inevitably increases regret. As a result, no single policy can simultaneously minimize both objectives.

Motivated by this intuition, we measure the performance of any sampling policy $\pi$ by a pair $(\mathrm{Regret}_T(\pi), \mathrm{RMSE}_T(\pi))$, where each term denotes the worst-case value of the corresponding objective:
\begin{align*}
    &\mathrm{Regret}_T(\pi) = \max_{(\A, \theta^*) \in \E} \mathrm{Regret}_T(\pi, \A, \theta^*), \\
    &\mathrm{RMSE}_T(\pi) = \max_{(\A, \theta^*) \in \E} \mathrm{RMSE}_T(\pi, \A, \theta^*).
\end{align*}


To compare policies, we characterize optimality in terms of \textit{Pareto domination}. A policy $\pi^*$ is Pareto-optimal if there exists no other policy $\pi$ such that
$$
\mathrm{Regret}_T(\pi) \le \mathrm{Regret}_T(\pi^*) 
\quad \text{and} \quad 
\mathrm{RMSE}_T(\pi) \le \mathrm{RMSE}_T(\pi^*),
$$
with at least one inequality strict. The set of all Pareto-optimal policies defines the \textit{Pareto frontier} ($
\min_{\pi} F(\pi) = \left[ \, \mathrm{Regret}_T(\pi),\, \mathrm{RMSE}_T(\pi) \,\right]
$), representing the fundamental boundary of achievable trade-offs: any improvement in one objective beyond this frontier necessarily worsens the other. Our goal is therefore not to find a single globally optimal policy but to design bandit algorithms that enable controlled movement along the Pareto frontier, allowing practitioners to select policies aligned with desired trade-offs that are application-specific.

\begin{pblm}{\textbf{Learning Peer Influence Probabilities under Bi-Objective Trade-off}}
Given an attributed network \( G = (V, E) \), the goal is to learn peer influence probabilities \( e_{ij}.p \) over \( T \) rounds by selecting top-$k$ edges in each round to minimize both regret and estimation error.
\end{pblm}

\section{Estimation-Regret Trade-Off}

In this section, we present theoretical results demonstrating the existence of a fundamental trade-off, in the worst case, between two objectives: minimizing cumulative regret and minimizing the RMSE of estimating the influence probabilities. We begin by establishing the achievable rates for each objective individually, thereby illustrating the inherent difficulty of both tasks. We then prove that there exist instances in which a trade-off arises between these two objectives, showing that no algorithm can achieve the optimal rate for both simultaneously. Finally, we propose an algorithm that, by tuning a parameter, can interpolate between these objectives and attain any achievable pair of rates implied by the lower bound.

Note that in the described problem setting, it is not possible to generally establish a worst-case bound on the objectives, particularly for the RMSE. The main difficulty arises because, at each round $t$, nature selects the set of available actions $\A_t$ (e.g., which specific users have taken an action to be shared with their friends), and the algorithm must then choose $k$ of its members to play. Since $\A_t$ is determined by nature and may vary across rounds, a uniform worst-case bound on the RMSE cannot be guaranteed. 
For example, suppose the actions can be partitioned into two sets $S_1$ and $S_2$ such that all vectors $X_m \in S_1$ are nearly collinear (i.e., point in almost the same direction), while the vectors in $S_2$ lie far from that direction. If nature selects $\A_t \subseteq S_1$ for most rounds, the estimation of the parameter vector $\theta^*$ will be poor in directions corresponding to $S_2$. Consequently, the estimation errors $| X_m^{\top}\hat{\theta}_T - X_m^{\top}\theta^* |$ will remain large for each $X_m \in S_2$, resulting in a high RMSE.

Motivated by this limitation, we establish our theoretical results in a slightly modified setting where nature cannot restrict the action set across rounds, and the agent has access to a fixed set of actions throughout. 
While our theoretical bounds and guarantees are derived for this fixed-action setting, our algorithm is set up (and our experiments conducted) under the more general settings introduced earlier.
Formally, for the theoretical analysis, we assume that the set of available actions at each round is constant, i.e., $\A_t = \A$ for all $t$. We now proceed to provide bounds for each objective under this setting to highlight their inherent difficulty.

For the regret minimization objective, the agent may choose any set of $k$ actions from $\A$ and then observe binary rewards indicating whether activation is successful for each of the selected arms. This setup corresponds to the well-studied \textit{combinatorial linear bandit} problem with semi-bandit feedback~\cite{CB1-cesa2012combinatorial, comblinucb-wen2015efficient, comblin-lb-audibert2014regret, LCB-combes2015combinatorial}. For this problem, the \textsc{CombLinUCB} algorithm introduced in~\cite{comblinucb-wen2015efficient} achieves the following worst-case regret upper bound over $T$ rounds: $\tilde{\mathcal{O}}\!\left(k d \sqrt{T}\right)$
where $\tilde{\mathcal{O}}$ hides constant and logarithmic factors.  
Regarding tightness, a lower bound of $\Omega\!\left(d\sqrt{kT}\right)$ was established in~\cite{takemura-aaai-2021}. Although these bounds do not match, they are close and can capture the inherent difficulty of the regret minimization problem.

For the objective of minimizing RMSE, we propose an algorithm and prove a worst-case upper bound on its RMSE over $T$ rounds. 
At each round $t$, the agent selects the edges for activation, together with their feature vectors $\{ X_{t,1}, X_{t,2}, \ldots, X_{t,k} \}$.
We define 
\begin{align} \label{eq: V_t}
V_t = \lambda \mathbb{I} + \sum_{s = 1}^{t} \sum_{r=1}^k X_{s,r} X_{s,r}^{\top},
\end{align}
where $\lambda > 0$ is a regularization constant. It is well known in the linear bandit literature that, for each $X_m \in \A$, the quantity $\sqrt{X_m^{\top} V_t^{-1} X_m}$ would characterize the estimation uncertainty of the influence probability associated with feature vector $X_i$ after $t$ rounds \cite{bandit-book-lattimore2020bandit}. Motivated by this intuition, to minimize the maximum estimation uncertainty over all actions, the classical linear bandit problem (i.e., for $k=1$) relies on the $G$-optimal design, which is obtained as the solution to the following optimization problem:
\begin{align} \label{eq: g-optimal}
    \min_{\wb \in \Delta^{M-1}} \max_{X_i \in \A} 
    X_i^{\top} \!\left( \sum_{j=1}^M w_j X_j X_j^{\top} \right)^{-1} \!X_i,
\end{align}
where $\Delta^{M-1}$ denotes the $(M-1)$-dimensional simplex. In this setting, given the optimal solution $\wb^*$, the algorithm that plays each arm $X_i$ for $T w_i^*$ rounds minimizes the maximum estimation error of mean rewards.  

Our setting can be viewed as a linear bandit problem with $kT$ total plays. The key difference is that at each round, the agent must choose $k$ distinct arms, meaning no arm can be played more than $T$ times out of the total $kT$ selections. Consequently, if we were to adopt the $G$-optimal design, some coordinates of the optimal design vector $\wb^*$ might exceed $\tfrac{1}{k}$, which is infeasible in our setting. To address this, we consider a constrained design that minimizes the maximum uncertainty while ensuring that no arm is selected more than $T$ times. Formally, we define the optimization problem
\begin{align} \label{eq: gk-optimal}
    f^*(\A, k) = \min_{\wb \in \Delta^{M-1}_{1/k}} \max_{X_i \in \A} 
    X_i^{\top} \!\left( \sum_{j=1}^M w_j X_j X_j^{\top} \right)^{-1} \!X_i,
\end{align}
where 
$\Delta^{M-1}_{1/k} = \bigl\{ \wb \in \Delta^{M-1} \;\mid\; \forall i: w_i \leq \tfrac{1}{k} \bigr\}$
is the subset of the simplex in which all entries are bounded above by $\tfrac{1}{k}$.

\begin{lemma}[RMSE Upper Bound]\label{lem: rmse-upper}
For any instance $(\A, \theta^*) \in \E$, consider the algorithm that allocates its $kT$ selections proportionally to $\wb_k^*$, the optimizer of~\eqref{eq: gk-optimal}. Then the final (expected) RMSE satisfies
$$
\mathrm{RMSE}_T(\pi, \A, \theta^*)\in \tilde{\mathcal{O}}\!\left( \sqrt{\frac{f^*(\A, k)}{k \, T}} \right),
$$
where $\tilde{\mathcal{O}}(\cdot)$ hides logarithmic factors.
\end{lemma}

\noindent
All the theoretical proofs are provided in the 
Appendix \ref{apd: proofs}.
\noindent
In the special case where the optimal solution of the unconstrained problem~\eqref{eq: g-optimal} already lies in $\Delta^{M-1}_{1/k}$ and $\mathrm{span}(X_1,\ldots,X_M)=\mathbb{R}^d$, we have $f^*(\A,k)=d$; otherwise, $f^*(\A,k)$ is larger.

The bounds above imply that the optimal rates in $T$ for cumulative regret and RMSE are $T^{\frac{1}{2}}$ and $T^{-\frac{1}{2}}$, respectively. We now show that there exist instances in which these two objectives conflict and no algorithm can achieve both optimal rates simultaneously. 

\begin{theorem}[RMSE--Regret Trade-Off]\label{them: trade-off}
For any $M > k$, there exists an instance $(\A, \theta^*)$ such that, for any policy $\pi$ with
$$
\mathrm{Regret}_T\!\left(\pi, \A, \theta^*) \right) \in \mathcal{O}\!\left( T^{2\beta} \right),
$$
we have
$$
\mathrm{RMSE}_T\!\left(\pi, \A, \theta^*) \right) \in \Omega\!\left( T^{-\beta} \right),
$$
where $\beta \le \tfrac{1}{2}$, and by the regret lower bound, $\tfrac{1}{4} \le \beta$.
\end{theorem}

\begin{remark}[Dependence on $d$ and $k$]
\label{rem:dk}
Our trade-off analysis focuses intentionally on the dependence in $T$, since the main theoretical goal is to show that regret minimization and global estimation cannot, in general, be optimized simultaneously, and to characterize its dominant asymptotic behavior. The analysis does not characterize the dependence on other structural parameters, such as $d$ and $k$. 
\end{remark}

By Theorem~\ref{them: trade-off}, if an algorithm achieves the optimal rate for cumulative regret (i.e., $\beta=\tfrac{1}{4}$, giving $R_T=\mathcal{O}(T^{1/2})$), then its RMSE rate must be non-optimal, namely $\Omega\!\left(T^{-\tfrac{1}{4}}\right)$. Conversely, if an algorithm achieves the optimal RMSE rate (i.e., $\beta=\tfrac{1}{2}$, giving $\mathrm{RMSE}_T=\Theta(T^{-1/2})$), then its cumulative regret is necessarily linear. 
An important question is whether one can achieve any pair of rates satisfying the condition in Theorem~\ref{them: trade-off}. In the next section, we propose an algorithm that, based on the parameter $\beta$, attains any feasible pair of rates implied by the theorem.

\section{Influence Contextual Bandit framework}
We propose an \textbf{Influence} \textbf{C}ontextual \textbf{B}andit ($\Lhse$) framework that simultaneously learns peer influence probabilities and leverages them to recommend top-$k$ edges for intervention.  In each round, the framework chooses between uncertainty-guided exploration and reward-guided exploitation. The details of the two phases are presented next, with the pseudo-code provided in Algorithm~\ref{SpillCB}. Then we present a theorem on the optimality of $\Lhse$.

\subsection{Uncertainty-guided Exploration}
\label{first_component}
Contextual bandits face the well-known cold-start problem: in early rounds, limited feedback makes reward estimation highly unreliable, particularly in the setting of influence probability learning. To mitigate this, $\Lhse$ employs an uncertainty-guided exploration strategy. In each round, the framework computes the estimation uncertainty for all candidate edges from the action pool. It considers two parameters, $\beta$ and $C$ which determine different strategies for the tradeoff between regrets and RMSE. As discussed in the previous section, $\beta$ reflects our preference for prioritizing each of the objectives; for example, $\beta=1/4$ optimizes for regret, and $\beta=1/2$, for RMSE. The second parameter $C$ is learned and the learning procedure reflects whether we would like to achieve a tradeoff between regrets and RMSE or optimize fully for one of the objectives. While our theoretical analysis assumes a fixed threshold $C$, it is under the assumption of a fixed action pool over rounds. Instead, here we allow for optimizing $C$ under a more flexible setting.

If the maximum uncertainty across edges exceeds the threshold $C / t^{\beta}$ in a particular round $t$, the algorithm selects the exploration phase in that round. This adaptive thresholding mechanism ensures that actions with high uncertainty are explored sufficiently often, progressively reducing their uncertainty in subsequent rounds. As a result, the uncertainties of all actions remain at most on the order of $\mathcal{O}\!\big(T^{-\beta}\big)$, which leads to the corresponding upper bound on the RMSE. In this phase, the top-$k$ edges with the highest uncertainty scores are recommended for intervention. CMAB parameters are updated after the rewards for the selected edges are observed. By prioritizing
uncertain edges, the framework increases edge diversity and
accelerates the estimation of influence probabilities, albeit at the
cost of higher short-term regret.

\subsection{Reward-guided Exploitation} 
\label{second_component}
When the maximum edge uncertainty falls below the threshold $C/t^\beta$ in round $t$, $\Lhse$ is more confident in its predictions and selects the exploitation phase in that round. Given a fixed $\beta$, increasing $C$ biases the bandit towards exploitation (minimizing regret), leading to lower cumulative regret $Regret_T$ but higher estimation error $RMSE_T$. 
This step is performed by incorporating a linear CMAB algorithm designed for regret minimization, which is the input to our algorithm, and our algorithm plays based on its policy. We denote the sampling policy of the CMAB algorithm at round $t$ by $f(\mathcal{F}_{t-1})$, where $\mathcal{F}_{t-1}$ is the history of interaction until round $t-1$.
The rewards from the corresponding edge actions are observed, and the CMAB parameters are updated to refine subsequent reward estimates. By exploiting the learned knowledge, the framework gradually improves the quality of interventions while reducing regret over rounds.
\paragraph{Oracle exactness.}
The CombLinUCB regret guarantee invoked here requires an exact
combinatorial oracle. In our setting this oracle is trivial: feasible
actions are all size-$k$ subsets of the current pool $\mathcal{A}_t$,
and the objective is additive across selected edges, so maximizing the
UCB score reduces to selecting the $k$ edges with the largest UCB
values. The neighbor-based setting inherits the same property.

\begin{algorithm}[tb] 
\caption{Influence Contextual Bandit Framework} 
\label{SpillCB}
\begin{algorithmic}[1]
    \State \textbf{Input:} rounds $T$; CMAB hyperparameter $\alpha$; interventions per round $k$; exploration parameter $\beta$; objective $\mathcal{O} \in \{\text{Regret}, \text{RMSE}\}$, linear CMAB $f(\mathcal{F}_t)$.
    \State \textbf{Output:} Set of $k$ edges with features $\mathbf{X}_t = \{X_{t,1}, X_{t,2},\ldots,X_{t,k}\}$ to intervene on at each round $t$.
    \For{each round $t = 1, 2, \ldots, T$}
        \State Agent receives the pool of available actions $\A_t \in \A$.
        \State \textbf{for each} $X\in \A_t$: $U_t(X) \gets \sqrt{X^{\top} V_{t-1}^{-1}X}$, where $V_{t-1}$ is defined in \eqref{eq: V_t}.
        \State Set $u_t \gets \max_{X \in \A_t} U_t(X)$, $au_t \gets mean_{X \in \A_t} U_t(X)$.
            \State $C_t \leftarrow UpdateC(\mathcal{O}, au_t,  \text{activation rate})$
            \Comment{\textit{Algorithm~\ref{alg:updateC}}}

        \If{$u_t > C_t / t^{\beta}$} \Comment{\textit{Exploration (uncertainty-guided)}}
            \State $\mathbf{X}_t \leftarrow \underset{X \in \A_t}{\arg top_k}\; U_t(X)$
        \Else \Comment{\textit{Exploitation (reward-guided)}}
            \State $\mathbf{X}_t \leftarrow f(\mathcal{F}_{t-1})$
        \EndIf
        \State Observe rewards $r_{t,1}, r_{t,2},\ldots, r_{t,k}$, and update  $\hat\theta_t$ , $V_t$. 
    \EndFor
\end{algorithmic}
\end{algorithm}
\subsection{Illustration of the $\Lhse$ Algorithm}
Algorithm~\ref{SpillCB} shows the pseudo-code for $\Lhse$ which operates over discrete rounds \(t = 1, 2, \ldots, T\).
In each round \(t\), the following steps are executed:

\begin{enumerate}
    \item \textbf{Action Pool Observation.}  
    The agent receives the set of candidate edge contexts \(\A_t\) from the network, representing the pool of available actions for round \(t\) \textit{[Line 4]}.

    \item \textbf{Uncertainty Estimation.}  
    For each edge \(e_{ij}\) with feature vector \(X \in \A_t\), the algorithm computes the associated uncertainty \( U_{t}(X) = \sqrt{X^\top V_{t-1}^{-1} X} \) \textit{[Line 5]}.

    \item \textbf{Maximum Uncertainty Computation.}  
    Compute the maximum uncertainty \(u_t = \max_{X \in \A_t} U_t(X)\) to quantify the exploration potential of the current action pool \textit{[Line 6]}.

    \item \textbf{Threshold Computation.}  
    A dynamic threshold \(C_t\) is computed according to the optimization objective \(\mathcal{O}\) \textit{[Line 7]}:  
    \begin{itemize}
        \item If \(\mathcal{O} = \text{RMSE}\), \(C_t\) is derived from uncertainty statistics to encourage exploration (Algorithm~\ref{alg:updateC}).  
        \item If \(\mathcal{O} = \text{Regret}\), \(C_t\) is computed from activation statistics to bias exploitation (Algorithm~\ref{alg:updateC}).
    \end{itemize}
$C_t$ adapts over time based on recent batch-level statistics. In the uncertainty-based version, it decreases when current uncertainty exceeds its historical mean to encourage exploration and increases when it falls below the historical mean to favor exploitation. In the activation-based version, it increases when current activation rates drop below their historical mean, steering the model toward exploitation.
    \item \textbf{Exploration--Exploitation Decision.}  
    If \(u_t > C_t / t^{\beta}\), the algorithm enters the \textit{exploration phase}, selecting the top-\(k\) edges with the highest uncertainty scores. Otherwise, it enters the \textit{exploitation phase},
    where the algorithm plays by the policy of the input linear CMAB algorithm \textit{[Lines 8--12]} (e.g., CombLinUCB, CombLinTS \cite{comblinucb-wen2015efficient}, or any other algorithm with similar regret bounds).
    \item \textbf{Reward Observation and Parameter Update.}  
    After selecting the intervention set \(X_t\), the algorithm observes the rewards associated with the corresponding edges and model parameters are updated with the new feedback \textit{[Line 13]}.
\end{enumerate}

\begin{algorithm}[t]
\caption{UpdateC: Adaptive Computation of $C_t$}
\label{alg:updateC}
\begin{algorithmic}[1]
\State \textbf{Input:} objective $\mathcal{O} \in \{\text{RMSE}, \text{Regret}\}$, \text{activation rate}, \text{average uncertainty}, $C_{max}$, $C_{min}$, $\gamma$, $warmup$, $\epsilon$
\State $m_t \gets \text{activation rate}$ if $\mathcal{O} = Regret$ else $\text{average  uncertainty}$
\State Append $m_t$ to history buffer $hist$
\State $baseline \gets mean(hist)$ if $|hist| > warmup$ else $m_t$
\State $z \gets (baseline - m_t) / (std(hist) + \epsilon)$
\If{$\mathcal{O} = \text{Regret}$} $z \gets \max(0, z)$ \EndIf
\State $C_{new} \gets C_{min} + (C_{max} - C_{min}) / (1 + e^{-\gamma z})$
\If{$\mathcal{O} = \text{Regret}$} 
  \State $C_t \gets \max(C_{prev}, C_{new})$, $C_{prev} \gets C_t$
\Else 
  \State $C_t \gets C_{new}$
\EndIf
\State \textbf{return} $C_t$
\end{algorithmic}
\end{algorithm}

The computational complexity of $\Lhse$ is governed primarily by the size of \(\A_t\), the candidate action pool, rather than total graph size, with uncertainty computations scaling as $\mathcal{O}(|A_t|d^2)$. Its adaptive exploration--exploitation mechanism, modulated by \(\beta\) and \(C_t\), enables continuous navigation along the Pareto frontier, balancing regret minimization against estimation accuracy. 

The next theorem shows that, by choosing $\beta$, the algorithm attains all rates on the Pareto frontier (in $T$) implied by the lower bound in Theorem~\ref{them: trade-off}. 

\begin{theorem}[Achievability of the Pareto-Frontier]\label{thm: upper bound}
    For any $\beta \in \bigl[\tfrac{1}{4}, \tfrac{1}{2}\bigr]$, Algorithm~\ref{SpillCB} initialized with parameter $\beta$ on any instance $(\A, \theta^*)$ with $\forall t: \A_t = \A, C_t = C$, where $C$ is a positive constant, satisfies
    \begin{align*} 
    &\mathrm{Regret}_T\!\left(\text{InfluenceCB}, \A, \theta^* \right) \in \tilde{\mathcal{O}}\!\left(T^{2\beta}\right), \\
    &\mathrm{RMSE}_T\!\left(\text{InfluenceCB}, \A, \theta^* \right)  \in \tilde{\mathcal{O}}\!\left(T^{-\beta}\right),
    \end{align*}
    where $\tilde{\mathcal{O}}(\cdot)$ hides logarithmic factors.
\end{theorem}

\textbf{Proof Sketch.} Here, we provide a high-level overview of the proof with full proof in the Appendix. We first show that the uncertainty of any arm $X_m \in \A$ decreases at least on the order of $\tfrac{1}{\sqrt{N_t(X_m)}}$, where $N_t(X_m)$ denotes the number of times arm $X_m$ has been selected up to round $t$. 
Next, we show that, under the exploration condition of our algorithm, the total number of exploration rounds is bounded by $\mathcal{O}(T^{2\beta})$, which, combined with the regret bound of the regret minimization subroutine, implies the regret upper bound. 
Finally, we prove that the proposed exploration mechanism guarantees that the uncertainty of each arm remains at most on the order of $\mathcal{O}(T^{-\beta})$, yielding an estimation accuracy of the same order and establishing the RMSE upper bound. \qed 

Theorem~\ref{thm: upper bound} is stated for the fixed action set, fixed threshold setting, i.e., $\forall t:\,\A_t=\A$ and $C_t=C$, which is the regime for which our formal guarantees hold. In the experiments in Section~\ref{sec: experiment}, we also consider varying action pools $\A_t$ and propose an adaptive thresholding rule for $C_t$ as an empirical enhancement. We emphasize that this adaptive mechanism is used only as a practical tool and is not part of the core theoretical claim.
Still, since the adaptive rule keeps $C_t \in [C_{\min},C_{\max}]$ for all $t$, the proof of Theorem~\ref{thm: upper bound} extends by replacing $C$ with $C_{\min}$ in the regret argument and with $C_{\max}$ in the RMSE argument, yielding regret of order $\tilde{\mathcal{O}}(C_{\min}^{-2}T^{2\beta})$ and RMSE of order $\tilde{\mathcal{O}}(C_{\max}T^{-\beta})$. In particular, choosing $C=(C_{\min}^2C_{\max})^{1/3}$ equalizes the two multiplicative gaps, giving at most a factor $\left({C_{\max}}/{C_{\min}}\right)^{2/3}$ compared to the corresponding fixed-$C$ bound. Thus, adaptivity affects only the constant factors and not the rates in $T$, but a full theoretical analysis of the adaptive and varying pool setting is left for future work.



\section{Experiments} \label{sec: experiment}
We evaluate the performance of $\Lhse$ on semi-synthetic network datasets against static, online, and bandit baselines, demonstrating its advantages and revealing the trade-off between minimizing regret and minimizing estimation error. 
\subsection{Data representation}
We generate ground-truth peer influence probabilities for three real-world attributed network datasets.
BlogCatalog~\cite{yang13homogeneous} 
is a social network, consisting of $5{,}196$ bloggers, $343{,}486$ following relationships, and $8{,}189$ blogger attributes.
Flickr~\cite{yang13homogeneous} is a photo-sharing network of $7{,}575$ users, $479{,}476$ following relationships, and $12{,}047$ user attributes. 
Hateful~\cite{ribeiro-aaai-2018} is a Twitter network with $100{,}386$ users, $2{,}194{,}979$ edges, and $1{,}036$ user attributes where there are directed edges between every user and anyone who has retweeted them. 

\subsubsection{Construction of edge feature representations}
\label{edge_features}
The feature representation of an edge $e_{ij} \in E$ is constructed by 
concatenating the feature vectors of its endpoints $v_i$ and $v_j$, and their network information.
To represent the attributes of each node, we apply 
truncated SVD~\cite{halko-book-2009} and project its attributes into a $64$-dimensional space. To capture structural properties, we generate $64$-dimensional node embeddings using a two-layer GraphSAGE model with mean aggregation~\cite{hamilton-neurips-2017}. The final edge vector has 256 dimensions.

\subsubsection{Generation of synthetic ground truth peer influence probabilities (linear).}
\label{section-spillover-generation-linear}
The peer influence probability associated with edge $e_{ij}$ is synthetically generated as a linear combination of the features, $e_{ij}.p = \sigma\!\left( \mathbf{w}^\top e_{ij}.X \right)$ (justification provided
in the 
Appendix~\ref{section-justify-linear-model}), 
where $\mathbf{w} \in \mathbb{R}^{256}$ is a weight vector with entries sampled 
uniformly at random from $[-1,1]$, and $\sigma(\cdot)$ is a min-max scaling function 
ensuring $e_{ij}.p \in (0,1)$. For our sensitivity analysis, we also considered a misspecified non-linear model (described in the 
Appendix~\ref{section-non-linear}).

\subsection{Evaluation Metrics}
We evaluate $\Lhse$ using four complementary metrics.
$Regret_T$ measures cumulative reward loss relative to the optimal top-$k$ edge
selection over $T$ rounds under ground-truth influence probabilities.
$RMSE_T$ evaluates the accuracy of learned influence probabilities over a random held-out set of edges after $T$ rounds. We also report on Expected Calibration Error ($ECE_T$) which assesses the calibration of predicted influence probabilities after $T$ rounds, and on $NDCG@k$ which measures the ranking quality of the selected top-$k$ edges with respect to the true influence ordering over $T$ rounds. Formal definitions of all metrics are provided in the 
Appendix~\ref{section-detailed-metrics}.

\subsection{Main Algorithms and Baselines}
\subsubsection{\textbf{Static Models}} 
We consider several representative static baselines that do not learn peer influence probabilities over rounds. 

\textit{Random.} Each network edge is pre-assigned a random score sampled uniformly from $[0,1]$. In every round, this baseline selects $k$ edges with the highest random scores from the candidate pool.

\textit{Similarity.} Following~\cite{jiang-arxiv-2015}, at each round this method selects the $k$ edges with the smallest Euclidean distance between the feature vectors of their two endpoint nodes.

\textit{Link Prediction (Ridge).} This baseline adopts a supervised learning approach to predict edge weights, following the link prediction framework in~\cite{kumar-pa-2020}. Existing edges $E$ are labeled as $1$, while an equal number of randomly sampled non-existing edges are labeled as $0$. 
A ridge regression model is trained on this labeled data to estimate link scores for all edges $e_{ij}\in E$, used for selecting the top k edges. 

\textit{Link Prediction (GraphSAGE).} This method is similar to the Ridge baseline except that it leverages the GraphSAGE algorithm~\cite{hamilton-neurips-2017}, following the hyperparameter setting of ~\cite{lu-neurocomputing-2024}. 

\textit{Link Prediction (NCNC).} This baseline employs the Neural Common Neighbor (NCNC) model~\cite{wang-iclr-2024}, a state-of-the-art link prediction approach to generate the link score of all edges in the network, which is then used to rank edges at each round. 
We follow the default configuration for hyperparameter setting~\cite{wang-iclr-2024}.

\textit{SOTA Recommender (ContextGNN).}
We adapt a state-of-the-art graph recommender, ContextGNN~\cite{yuan-iclr-2025}, by converting the user–user network into a bipartite user–item interaction graph via node duplication, where an interaction edge from user $u$ to item $v$ is introduced if $v$ is in the $1$-hop neighborhood of $u$ in the original network. The model is trained offline following ~\cite{yuan-iclr-2025} and is evaluated as a static scoring baseline; in each round, the top-$k$ candidate edges are selected according to the ContextGNN scores. 

\subsubsection{\textbf{Online Learning Models}}  
We employ an online learning baseline in which model parameters are updated incrementally using Stochastic Gradient Descent (SGD)~\cite{bottou-compstat-2010}. Each of the $k$ selected edges triggers a gradient update, allowing the model to adapt continuously without retraining from scratch. Specifically, we adopt \texttt{SGDClassifier} with the \texttt{partial\_fit} routine~\cite{pedregosa-jmld-2011}, where each of the $k$ selected edges is labeled as positive if the recipient node becomes activated and negative otherwise. In each round, the top-k edges are selected based on their SGD-estimated probability of being positive. 
We consider two variants of this baseline:

\textit{$SGD_{Exploit}$.} A purely exploitative variant where, in each round, the top-$k$ edges with the highest estimated probabilities are selected.

\textit{$SGD_{Explore}$.} A purely exploratory variant where, in each round, $k$ edges are selected uniformly at random from the candidate pool. 

\subsubsection{\textbf{Bandit Models}}  
We also evaluate four CMAB algorithms: 

\textit{$LinUCB$.} This baseline assumes that the expected reward 
is a linear function of the edge feature vector~\cite{li-www-2010}. 
We set $\alpha = 2.0$.

\textit{$EENet$.} This is a nonlinear baseline~\cite{ban-iclr-2022} which 
can capture more complex feature–influence relationships. We follow the authors’ default configuration~\cite{ban-iclr-2022} and perform a grid search over learning rates for each dataset, reporting the best-performing results. 

\textit{GNB (Graph Neural Bandits).} A graph-neural-network-based
contextual bandit~\cite{qi-kdd-2023}. We use the authors' default
configuration.

\textit{Takemura et al.} A near-optimal contextual combinatorial
semi-bandit algorithm with tighter regret
bounds~\cite{takemura-aaai-2021}; we follow the authors' default
configuration.

\begin{figure*}[t]
    \centering
     \begin{subfigure}
         \centering \includegraphics[width=\textwidth]{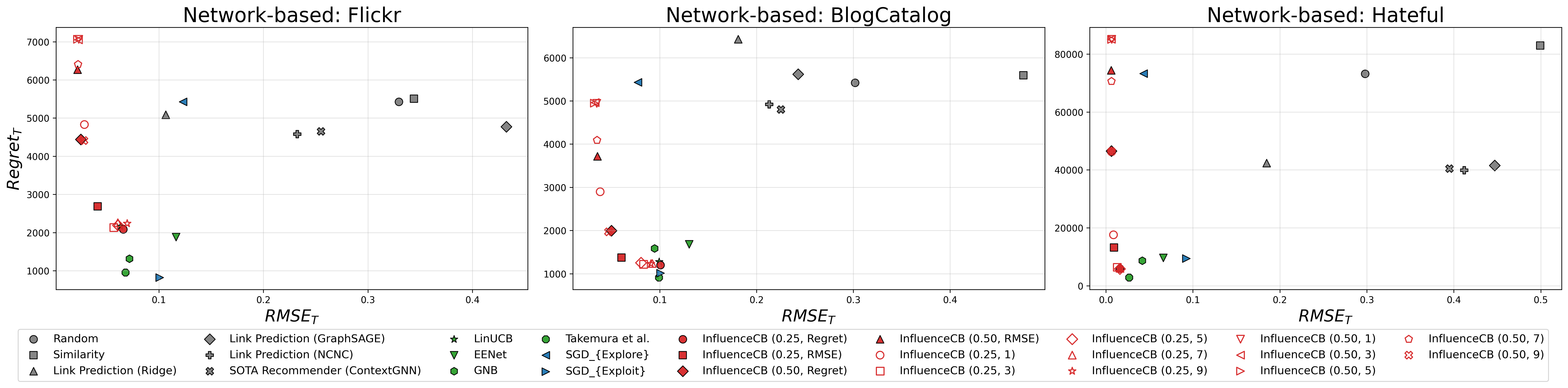}
     \end{subfigure}
     \vspace{-13 pt}
    \caption{Pareto frontier of the $Regret_T$--$RMSE_T$ tradeoff for $\mathsf{InfluenceCB}$ (solid red) and $\mathsf{InfluenceCB}$ with fixed C (red outline), compared to baseline static (grey), bandit (green), online (blue) models in network-based setting.}
     \label{fig-pareto-network-all}
\end{figure*}

\subsubsection{\textbf{Bandit Models for Influence ($\Lhse$)}}  
Our proposed framework, $\Lhse$, leverages CombLinUCB~\cite{comblinucb-wen2015efficient} as the CMAB algorithm to learn edge-level peer influence probabilities. It dynamically learns $C_t$ over rounds to optimize a specific objective $\mathcal{O}$ ($RMSE$ or $Regret$) as $\Lhse(\beta, \mathcal{O})$. When $C_t$ is fixed across rounds, we denote the variant as $\Lhse(\beta, C)$.

\subsection{Experimental setup} 
Our experimental setup constitutes a controlled counterfactual simulation with a fixed causal diffusion mechanism where the factual is realized based on the top-k choice of each algorithm and the counterfactual corresponds to the (unrealized) optimal top-k.
In each round, for each action, a reward is generated based on the true peer influence probability (unknown to the algorithms). To compute $RMSE_T$, we remove a fixed random subset of $|\overline{E}| = 500$ edges from the network to ensure an unbiased evaluation. We conduct experiments under two edge selection settings in each round: (i) selecting $k$ edges from a pool of $200$ edges randomly sampled from the network (\textit{network-based}), and (ii) selecting $k$ edges from the pool of a randomly selected node’s neighbors (\textit{neighbor-based}) in each round. The network-based experiments approximate the theoretical scenario where the fixed pool is all edges. Calculations for all edges at each round are very expensive which is why we limit the experiment to a random sample of $200$ edges per round. The neighbor-based experiments represent a realistic scenario where an agent’s action pool is localized to a specific node's immediate neighborhood. Due to space constraints, we include the results for the network-based setting in the main paper and the results for the neighbor-based setting in the 
Appendix
noting notable differences between them when needed. We consider $k \in \{5, 10, 15, 20, 25\}$. All experiments are repeated $10$ times with averages reported. 

Most of our experiments consider $\Lhse(\beta, \mathcal{O})$ with $\beta \in \{0.25, 0.50\}$. 
For our \emph{ablation studies}, we fix $C \in \{1, 3, 5, 7, 9\}$, rather than learning it dynamically, denoting this setting $\Lhse(\beta, C)$. We vary $\beta \in \{0.0, 0.25, 0.30, 0.35, 0.40, 0.45, 0.50, 0.75, 1.0\}$. We also remove the exploration component from $\Lhse(\beta, \mathcal{O})$ reducing it to LinUCB, to study the value of our uncertainty-guided exploration. Finally, we examine the sensitivity of our method to the choice of exploitation strategy by replacing LinUCB with another linear CMAB algorithm with similar regret guarantees, LinTS~\cite{agrawal-icml-2013}. The results are included in the 
Appendix~\ref{ablation-LinTS}.

\begin{table}[t]
\centering
\caption{$Regret_T$ and $RMSE_T$ achieved by $\Lhse(\beta, \mathcal{O})$ and baselines on the Flickr, BlogCatalog, and Hateful datasets with network-based edge selection (lower is better).}
\vspace{-8pt}
\label{table-network-based-half}
\resizebox{\columnwidth}{!}{
\begin{tabular}{lcccccc}
\toprule
\multirow{2}{*}{\textbf{Method}} 
& \multicolumn{2}{c}{\textbf{Flickr}} 
& \multicolumn{2}{c}{\textbf{BlogCatalog}} 
& \multicolumn{2}{c}{\textbf{Hateful}} \\
\cmidrule(lr){2-3} \cmidrule(lr){4-5} \cmidrule(lr){6-7}
 & $Regret_T$ & $RMSE_T$ & $Regret_T$ & $RMSE_T$ & $Regret_T$ & $RMSE_T$ \\
\midrule
\multicolumn{7}{c}{\textit{Static Models}} \\
\midrule
Random        & 5428 & 0.3296 & 5419 & 0.3018 & 73226 & 0.2981 \\
Similarity    & 5506 & 0.3440 & 5590 & 0.4757 & 82962 & 0.4993 \\
Link Prediction (Ridge)    & 5090 & 0.1066 & 6426 & 0.1810 & 42384 & 0.1848 \\
Link Prediction (GraphSAGE)& 4772 & 0.4325 & 5616 & 0.2430 & 41568 & 0.4470 \\
Link Prediction (NCNC)     & 4579 & 0.2325 & 4918 & 0.2130 & 39850 & 0.4120 \\
SOTA Recommender (ContextGNN)    & 4650 & 0.2550 & 4800 & 0.2250 & 40500 & 0.3950 \\
\midrule
\multicolumn{7}{c}{\textit{Bandit Models}} \\
\midrule
$LinUCB$ & 2165 & 0.0632 & 1275 & 0.0993 & 5979 & 0.0156 \\
$EENet$  & 1884 & 0.1164 & 1683 & 0.1302 & 9635 & 0.0658 \\
$GNB$  & 1314 & 0.0719 & 1586 & 0.0945 & 8669 & 0.0417 \\
$\textit{Takemura et al.}$  & 952 & 0.0681 & 912 & 0.0990 & 2829 & 0.0267 \\
\midrule
\multicolumn{7}{c}{\textit{Online Learning}} \\
\midrule
$SGD_{Explore}$ & 5425 & 0.1232 & 5432 & 0.0773 & 73297 & 0.0433 \\
$SGD_{Exploit}$ & 822  & 0.1006 & 1015 & 0.1007 & 9441  & 0.0924 \\
\midrule
\multicolumn{7}{c}{\textit{Ours: $\Lhse$}} \\
\midrule
$\Lhse (0.25, Regret)$ & 2083 & 0.0659 & 1202 & 0.1006 & 5917 & 0.0160 \\
$\Lhse (0.25, RMSE)$   & 2686 & 0.0414 & 1372 & 0.0605 & 13179 & 0.0092 \\
$\Lhse (0.50, Regret)$ & 4443 & 0.0254 & 1994 & 0.0500 & 46561 & 0.0062 \\
$\Lhse (0.50, RMSE)$   & 6271 & 0.0222 & 3719 & 0.0354 & 74399 & 0.0061 \\
\bottomrule
\end{tabular}
}
\vspace{-6pt}
\end{table}

\subsection{Experimental results}
\subsubsection{Comparing $\Lhse(\beta, \mathcal{O})$ to the baselines for $k=5$.}
Table~\ref{table-network-based-half} 
summarizes the performance of all methods for the network-based scenario. 
Figure~\ref{fig-pareto-network-all} shows the RMSE/regret tradeoff, including variants of $\Lhse(\beta, C)$ with fixed C. As expected, all static models (marked in grey) and $SGD_{Explore}$ achieve significantly worse RMSE and Regret tradeoffs than $\Lhse$. Interestingly, $SGD_{Exploit}$ sometimes achieves a better regret than our method which we attribute to our choice of exploitation strategy $LinUCB$ but it cannot achieve lower error.
In comparison to other bandit algorithms ($LinUCB$, $EENet$, $GNB$, and $\textit{Takemura et al.}$), we see that $\Lhse$ expands the Pareto frontier beyond them, providing tunable trade-offs between $Regret_T$ and $RMSE_T$ depending on the application's priorities. 

On Flickr, $\Lhse(0.25,\textit{Regret})$ achieves $3.8\%$ lower $Regret_T$ than $LinUCB$ with $4.3\%$ increase in $RMSE_T$. $\Lhse(0.25,\textit{RMSE})$ reduces $RMSE_T$ by $34.5\%$ at the cost of a $24.1\%$ increase in $Regret_T$. Increasing exploration ($\beta=0.5$) reduces $RMSE_T$ by $59.8\%$ and $64.9\%$ for $\Lhse(0.50,\textit{Regret})$ and $\Lhse(0.50,\textit{RMSE})$, respectively, but incurring higher $Regret_T$ ($105.2\%$ and $189.7\%$). A similar trade-off pattern is observed on BlogCatalog. $\Lhse(0.25, \\ \textit{Regret})$ outperforms $LinUCB$ in $Regret_T$ by $5.7\%$ with a slight ($1.3\%$) $RMSE_T$ increase, while $\Lhse(0.25,\textit{RMSE})$ achieves a $39.1\%$ reduction in $RMSE_T$ with a $7.6\%$ $Regret_T$ increase. High-$\beta$ configurations further improve estimation accuracy, with \\$\Lhse(0.50,\textit{Regret})$ and $\Lhse(0.50,\textit{RMSE})$ lowering $RMSE_T$ by $49.6\%$ and $64.4\%$ at the cost of $56.4\%$ and $191.7\%$ higher $Regret_T$, respectively. On the Hateful dataset, $\Lhse(0.25,\textit{Regret})$ yields a $1.0\%$ reduction in $Regret_T$ over $LinUCB$, accompanied by a $2.6\%$ increase in $RMSE_T$, whereas $\Lhse(0.25,\textit{RMSE})$ improves estimation accuracy with a $41.0\%$ reduction in $RMSE_T$ at the cost of a $120.4\%$ increase in $Regret_T$. Higher $\beta$ further reduces estimation error $RMSE_T$ by $60.3\%$ for $\Lhse(0.50,\textit{Regret})$ and $60.9\%$ for $\Lhse(0.50,\textit{RMSE})$, but incurring sharply higher $Regret_T$ ($678.7\%$ and $1144.3\%$). 
The neighbor-based experiments have a similar pattern.
We observe similar trends for the calibration error ($ECE_T$) and ranking metrics ($NDCG@k$) across all datasets.
$\Lhse$ improves $ECE_T$ relative to baselines,
especially under high-$\beta$ configurations. At the same time, $\Lhse$ maintains competitive
$NDCG@k$. Full results are included in the 
Appendix.


\subsubsection{Effect of $k$ and $d$ on the Trade-off Between $Regret_T$ and $RMSE_T$ in $\Lhse(\beta, \mathcal{O})$}  
Figure~\ref{fig:vary-k-heatmap} in the Appendix
shows how the number of interventions $k$ affects the increase in $Regret_T$ and reduction in $RMSE_T$ for $\Lhse(\beta, \mathcal{O})$ compared to $LinUCB$ in the network-based setting across the datasets. 
When RMSE is prioritized ($\beta=0.5$), then the estimation error reductions become smaller with increasing $k$ accompanied by larger regret increases, showing that the tradeoffs are best for smaller $k$. At the same time, when low regret is prioritized ($\beta=0.25$), the RMSE/regret tradeoff is not very sensitive to increase in $k$. 
This highlights $\Lhse$'s adaptability to varying $k$, enabling practitioners to balance exploration and exploitation according to deployment needs. Notably, the trends in the neighbor-based setting follow the same qualitative pattern but with more volatile changes in both $Regret_T$ and $RMSE_T$, indicating that the sensitivity to $k$ is stronger. Full results on the sensitivity to $d$ are included in the 
Appendix~\ref{sec:effect-of-d}.



\begin{figure}[t]
     \centering \includegraphics[width=0.95\columnwidth]{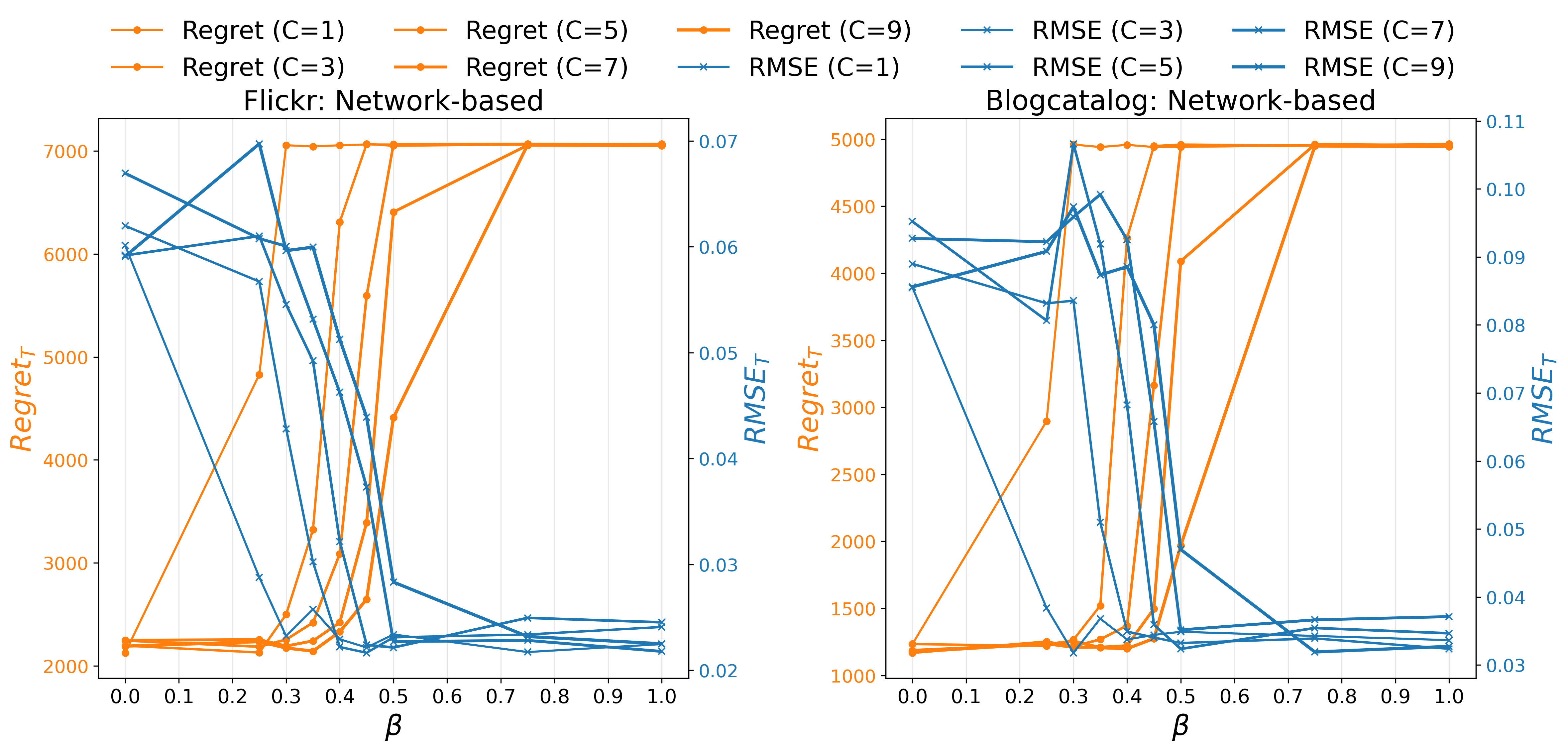}
    \caption{Trade-off between $Regret_T$ and $RMSE_T$ in $\Lhse(\beta, C)$ across network-based settings. Increasing $C$ consistently reduces $Regret_T$ but leads to higher $RMSE_T$, with the trade-off depending on the choice of $\beta$.}
     \label{fig-linear_C_network}
     \vspace{-10pt}
\end{figure}

\subsubsection{Ablation study}
\label{exp-1}
As we saw in Figure~\ref{fig-pareto-network-all}, fixing $C$ in the model allows us to explore the full Pareto frontier of  $Regret_T$ and $RMSE_T$ tradeoff. However, the optimal $C$ is not known in advance which is why $\Lhse$ learns it.
Figure~\ref{fig-linear_C_network} shows the trade-off between $Regret_T$ and $RMSE_T$ for different $C$,  demonstrating that given $\beta$, careful tuning of $C$ is crucial for achieving a good $Regret_T$--$RMSE_T$ tradeoff.  
When $\beta = 0.25$ (favoring optimal regret), increasing $C$ from $1$ to $9$ reduces $Regret_T$ (as expected) by $53.7\%$ and $57.5\%$ on the Flickr and BlogCatalog datasets, respectively, while increasing $RMSE_T$ by $142.1\%$ and $136.7\%$. Similarly, for $\beta = 0.50$ (favoring optimal RMSE), the same increase in $C$ results in a $37.6\%$ and $60.2\%$ reduction in $Regret_T$ on Flickr and BlogCatalog, respectively, accompanied by a $21.2\%$ and $34.7\%$ increase in $RMSE_T$. A similar pattern is observed in the neighbor-based experiments.

Furthermore, the choice of $C$ strongly influences the trade-off between regret minimization and estimation accuracy. When $\beta = 0.25$, $C = 3$ yields the lowest $Regret_T$, while $C = 1$ achieves the lowest $RMSE_T$ across both datasets. For $\beta = 0.50$, $C = 9$ leads to lowest $Regret_T$, whereas $C = 5$ leads to the lowest $RMSE_T$. Overall, $C = 3$ provides a good balance between $Regret_T$ and $RMSE_T$ for $\beta = 0.25$, whereas $C = 7$ achieves a good trade-off for $\beta = 0.50$. 

\section{Conclusion and Future Work}
This work introduced a novel formulation of learning heterogeneous peer influence probabilities in networked environments under a contextual linear bandit framework. We proved the existence of a fundamental trade-off between cumulative regret and estimation error, characterized the achievable rate pairs, and showed that no algorithm can achieve the optimal rates for both simultaneously. Building on this theoretical insight, we proposed $\Lhse$ that leverages uncertainty-guided exploration to navigate the Pareto frontier and enables explicit control over the exploration--exploitation balance through a parameter. Our empirical evaluation on semi-synthetic network datasets corroborates the theoretical results, demonstrating that $\Lhse$ can achieve good trade-off between estimation accuracy and regret, unlike static, online learning, and contextual bandit baselines. Our analysis assumes independent Bernoulli neighbor activations and does not model more complex interference effects such as cumulative exposures from multiple neighbors; extending the framework and guarantees to such complex network interference is an important direction for future work. Other future directions include studying tradeoffs in non-linear settings and considering dynamic or partially observed networks.




\bibliographystyle{ACM-Reference-Format}
\balance
\bibliography{aaai25}
\appendix
\section{Omitted Proofs} \label{apd: proofs}
In this section, we present the omitted proofs of the theoretical results. 

\subsection{Proof of Lemma~\ref{lem: rmse-upper}} 
Here, we present the proof of Lemma \ref{lem: rmse-upper}. 

\begin{proof}
    We first introduce two notations:
    \begin{align*}
        &\forall \wb \in \Delta^{M-1}_{1/k}: \quad 
        M(\wb)\;=\;\sum_{j=1}^M w_j X_j X_j^\top, \\ 
        &\forall \mathbf{u} \in \mathbb{R}^d, \; V \in \mathbb{R}^{d \times d}: \quad 
        \|\mathbf{u}\|_{V} = \sqrt{\mathbf{u}^{\top} V \mathbf{u}}. 
    \end{align*}
    If the $kT$ total selections are allocated proportionally to $\wb_k^*$, then the design matrix satisfies
    $$
    V_T \;=\; \lambda \mathbb{I} + \sum_{t=1}^{T} \sum_{r=1}^k  X_{t,r}X_{t,r}^\top 
    \;=\; \lambda \mathbb{I} + kT\,M(\wb_k^*).
    $$
    Since $A \succeq B \Rightarrow A^{-1}\preceq B^{-1}$, for any $X_i\in\A$,
    \begin{align*}
        &X_i^\top V_T^{-1}X_i 
        = X_i^\top\!\big(\lambda \mathbb{I} + kT\,M(\wb_k^*)\big)^{-1}\!X_i  \\
        &\le \frac{1}{kT}\,X_i^\top M(\wb_k^*)^{-1}X_i
        \;\le\;\frac{f^*(\A,k)}{kT}.
    \end{align*}
        
    Let $\hat\theta_T$ be the ridge estimator with parameter $\lambda>0$. Since rewards are binary, they are $\tfrac12$-sub-Gaussian. From the self-normalized bound in~\cite{abbasi2011improved}, for any $\delta \in (0,1)$, with probability at least $1-\delta$,
    $$
    \|\hat\theta_T-\theta^*\|_{V_T}
    \;\le\;\beta_T(\delta),
    $$
    where
    \begin{align} \label{eq: beta_T delta}
    \beta_T(\delta)\;=\;\tfrac12\sqrt{d\log\!\Big(\tfrac{1 + kT L^2/\lambda}{d}\Big)}\;+\;\sqrt{\lambda}\,\|\theta^*\|_2,
    \end{align}
    and $L = \max_i \|X_i\|_2$.
    
    To move from parameter error to prediction error for $X_i$, we apply Cauchy--Schwarz in the $V_T$-inner product:
    $$
    |X_i^\top(\hat\theta_T-\theta^*)|
    = \big| \langle V_T^{-1/2}X_i,\;V_T^{1/2}(\hat\theta_T-\theta^*) \rangle \big|
    \;\le\; \|X_i\|_{V_T^{-1}} \;\|\hat\theta_T-\theta^*\|_{V_T}.
    $$
    Thus, with probability at least $1-\delta$,
    $$
    |X_i^\top(\hat\theta_T-\theta^*)|
    \;\le\;\beta_T(\delta)\,\sqrt{X_i^\top V_T^{-1}X_i}
    \;\le\;\beta_T(\delta)\,\sqrt{\tfrac{f^*(\A,k)}{kT}}.
    $$
    
    By definition,
    $$
    \mathrm{RMSE}_T
    = \mathbb{E} \left[ \sqrt{\frac{1}{M}\sum_{i=1}^M \bigl(X_i^\top(\hat\theta_T-\theta^*)\bigr)^2} \right].
    $$
    The uniform bound above implies
    $$
    \mathrm{RMSE}_T
    \;\le\;
    \beta_T(\delta)\,\sqrt{\tfrac{f^*(\A,k)}{kT}}.
    $$
    
    Finally, choosing $\delta=\tfrac{1}{kT}$ and taking expectations yields
    $$
    \mathbb{E}\!\left[\mathrm{RMSE}_T\right]
    \;\le\;
    \tilde{\mathcal{O}}\!\left(\sqrt{\frac{f^*(\A,k)}{kT}}\right),
    $$
    where $\tilde{\mathcal{O}}(\cdot)$ hides logarithmic factors in $d$, $kT$, and confidence parameters. This completes the proof.
\end{proof}

\subsection{Proof of Theorem~\ref{them: trade-off}} 

\begin{proof}
We begin by presenting a known lower bound in the literature on the expected estimation error for Bernoulli random variables.

\begin{lemma}[Expected estimation error for the Bernoulli mean]
\label{lem: bernoulli-mean-LB}
Let $X_1, \dots, X_t$ be i.i.d.\ samples from a Bernoulli distribution with success probability $p \in [\tfrac{1}{4}, \tfrac{3}{4}]$, and let $\hat{p}$ be any estimator of $p$. Then there exists a universal constant $C > 0$ such that, for all $t \ge 1$ and all $p \in [\tfrac{1}{4}, \tfrac{3}{4}]$,
$$
\mathbb{E}_p\!\left[\,|\hat{p} - p|\,\right] 
\;\ge\; C\,\sqrt{\frac{p(1-p)}{t}}.
$$
\end{lemma}

We construct an instance $(\A_0, \theta_0^*)$ as follows.  
Let the action set be
$$
\A_0 = \{X_1, \dots, X_M\}, \quad
X_m =
\begin{cases}
(1,0,x_{m,3},\dots,x_{m,d}), & 1 \le m \le k,\\
(0,1,x_{m,3},\dots,x_{m,d}), & k < m \le M,
\end{cases}
$$
and set the true parameter vector to
$$
\theta_0^* = \Bigl(\tfrac{3}{4},\, \tfrac{1}{4},\, 0, \dots, 0\Bigr).
$$
The expected reward of arm $m$ is $X_m^\top \theta_0^*$, so the first $k$ arms are high-reward arms with mean $3/4$, and the remaining $M-k$ arms are low-reward arms with mean $1/4$.  
The expected reward gap is thus $1/2$.

Fix any policy $\pi$ that selects $k$ arms per round for $T$ rounds, for a total of $kT$ plays.  
Let $\alpha \in [0,1]$ denote the expected fraction of plays allocated to high-reward arms, so that a fraction $1-\alpha$ corresponds to plays of low-reward arms.  
Define $\beta = \frac{\ln(kT(1-\alpha))}{2 \ln T}$, equivalently $kT(1-\alpha) = T^{2\beta}$.  
The total expected regret is then
\begin{align*}
\mathrm{Regret}_T(\pi, \A_0, \theta_0^*)
&= \frac{1}{2} \times \bigl(\text{\# of plays on low-reward arms}\bigr)\\
&= \frac{1}{2} (kT)(1-\alpha) 
\;\in\; \mathcal{O}(T^{2\beta}).
\end{align*}

Let $N_{\mathrm{low}}$ denote the (random) number of plays of low-reward arms, so that $\mathbb{E}[N_{\mathrm{low}}] = kT(1-\alpha)$.  
Each time a low-reward arm is played, the observed reward follows $\mathrm{Bernoulli}(1/4)$ and provides information only about the second coordinate $\theta^*_{0,2} = \tfrac{1}{4}$.  
Conditioned on the (possibly adaptive) history of actions, these $N_{\mathrm{low}}$ samples are i.i.d.\ $\mathrm{Bernoulli}(1/4)$.

By the definition of RMSE,
\begin{align*}
\mathrm{RMSE}_T(\pi, \A_0, \theta_0^*)
&= \mathbb{E}\!\left[\sqrt{\frac{1}{|\A|}\!
    \sum_{X \in \A}\! \big(X^{\top} (\hat{\theta}_T - \theta^*_0)\big)^2}\right] \\
&\geq \mathbb{E}\!\left[\sqrt{\frac{1}{|\A|}\!
    \sum_{X \text{ low-reward}}\! \big(X^{\top} (\hat{\theta}_T - \theta^*_0)\big)^2}\right] \\
&\geq \sqrt{\frac{M-k}{M}} \; \mathbb{E}\!\left[|\hat{\theta}_{T,2} - \tfrac{1}{4}|\right],
\end{align*}
where $\hat{\theta}_{T,2}$ denotes the estimate of the second coordinate of $\theta^*_0$, and the last inequality follows from the Cauchy--Schwarz inequality.

Applying Lemma~\ref{lem: bernoulli-mean-LB} with $p = \tfrac{1}{4}$ yields
$$
\mathbb{E}\!\left[\,\bigl|\hat{\theta}_{T,2} - \tfrac{1}{4}\bigr| \,\bigm|\, N_{\mathrm{low}} = n\right]
\;\ge\; C'\,\frac{1}{\sqrt{n}},
\quad \text{for all } n \ge 1,
$$
for some universal constant $C' > 0$.  
Taking expectations and applying Jensen’s inequality to the convex function $n \mapsto 1/\sqrt{n}$, we obtain
$$
\mathbb{E}\!\left[|\hat{\theta}_{T,2} - \tfrac{1}{4}|\right]
\;\ge\; \frac{C'}{\sqrt{\mathbb{E}[N_{\mathrm{low}}]}}
\;=\; \frac{C'}{\sqrt{kT(1-\alpha)}}.
$$

Combining these inequalities gives
$$
\mathrm{RMSE}_T(\pi, \A_0, \theta_0^*)
\;\ge\;
C' \sqrt{\frac{M-k}{M kT(1-\alpha)}}
\;\in\;
\Omega(T^{-\beta}),
$$
where constants depending on $k$ are absorbed into the asymptotic notation.
This completes the proof.
\end{proof}

\begin{figure*}[t]
    \centering
     \begin{subfigure}
         \centering \includegraphics[width=\textwidth]{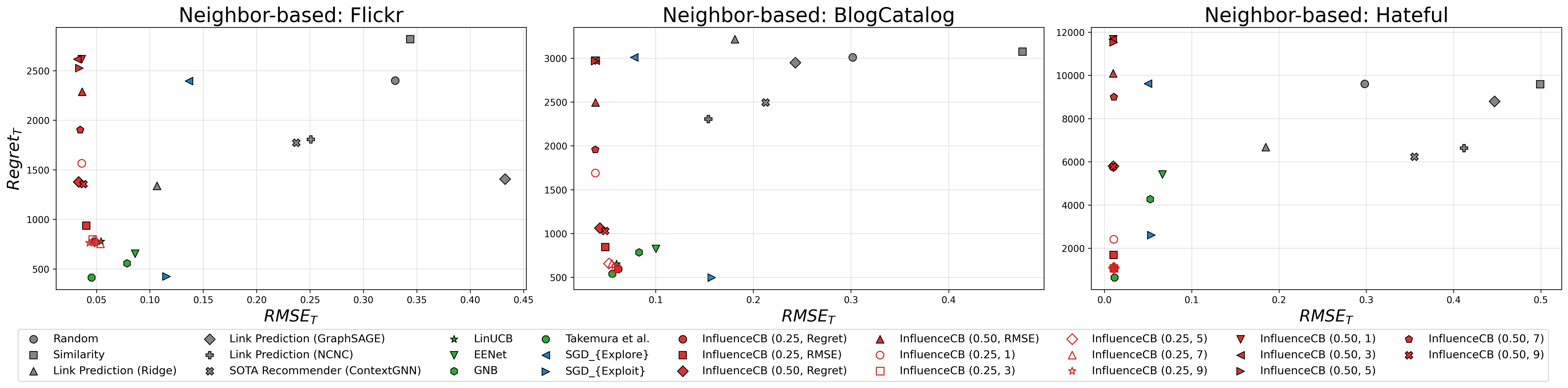}
     \end{subfigure}
     \vspace{-13 pt}
    \caption{Pareto frontier of cumulative regret $Regret_T$ and estimation error $RMSE_T$ tradeoff for $\mathsf{InfluenceCB}$ (solid red) and $\mathsf{InfluenceCB}$ with fixed C (red outline), compared to baseline static (grey), bandit (green), online (blue) models on semi-synthetic datasets in neighbor-based setting.}
     \label{fig-pareto-neighbor}
\end{figure*}

\subsection{Proof of Theorem~\ref{thm: upper bound}}

\begin{proof}

    Let the set of available actions be fixed as $\A$ for all rounds, and let $\forall t:\, C_t = C$.  
    For each round $t$ and arm $X \in \A$, denote by $N_t(X)$ the number of times Algorithm~\ref{SpillCB} has selected arm $X$ up to round $t$.  
    For any $\mathbf{u} \in \mathbb{R}^d$ and $V \in \mathbb{R}^{d \times d}$, define
    $$
    \|\mathbf{u}\|_{V} = \sqrt{\mathbf{u}^{\top} V \mathbf{u}}.
    $$
    The uncertainty of arm $X$ at round $t$ is given by $U_t(X) = \|X\|_{V_{t-1}^{-1}}$.  
    Let $m_t(X)$ denote the number of times $s \le t$ such that $u_s > C/s^{\beta}$ and $\arg\max_{X' \in \A} \|X'\|_{V_{s-1}^{-1}} = X$.

    We first show that for all $t$ and $X \in \A$,
    \begin{align} \label{eq: normvt upper bound by nt}
        \|X\|_{V_t^{-1}} \le \frac{1}{\sqrt{N_t(X)}}.
    \end{align}
    Recall
    \begin{align*}
    &V_t \;=\; \lambda \mathbb{I} \;+\; \sum_{s=1}^{t}\sum_{r=1}^k X_{s,r}X_{s,r}^\top \\
    &\;=\; \lambda \mathbb{I} \;+\; \sum_{X'\in\A} N_t(X')\,X'X'^\top 
    \;\succeq\; N_t(X)\,XX^\top.
    \end{align*}
    For any positive semidefinite matrix $A$ and any vector $u$, the \\ Cauchy–Schwarz inequality in the $A$-inner product yields
    $$
    u^\top A^{-1}u \;\le\; \frac{(u^\top u)^2}{u^\top A\,u}.
    $$
    Apply this with $A=V_t$ and $u=X$:
    \begin{align*}
    X^\top V_t^{-1} X 
    &\;\le\; \frac{\|X\|_2^4}{X^\top V_t X}
    \;\le\; \frac{\|X\|_2^4}{X^\top \bigl(N_t(X)\,XX^\top\bigr) X} \\
    &\;=\; \frac{\|X\|_2^4}{N_t(X)\,(X^\top X)^2}
    \;=\; \frac{1}{N_t(X)}.
    \end{align*}
    Equivalently,
    $$
    \|X\|_{V_t^{-1}} \;=\; \sqrt{X^\top V_t^{-1}X} \;\le\; \frac{1}{\sqrt{N_t(X)}}.
    $$
    
    Then, we prove the following upper bound on $m_t(X)$: 
    \begin{align} \label{eq: mt upper bound}
        m_t(X) \leq \frac{2t^{2 \beta}}{C^2}.
    \end{align}
    
    Let $s_1<s_2<\dots<s_{m_t(X)}$ be the rounds up to $t$ at which $X$ is (i) the maximizer of uncertainty and (ii) has uncertainty above the threshold, i.e.,
    $$
    \arg\max_{X'\in\A}\|X'\|_{V_{s_j-1}^{-1}} = X,
    \qquad
    \|X\|_{V_{s_j-1}^{-1}} > \frac{C}{s_j^\beta}.
    $$
    By the algorithm’s exploration rule, $X$ is selected at each such round, hence the play count increases strictly:
    $$
    N_{s_j}(X) \;\ge\; N_{s_j-1}(X)+1.
    $$
    Inductively, $N_{s_j-1}(X)\,\ge\, j-1$ for all $j\ge 1$.  
    Therefore, by Step~1,
    $$
    \|X\|_{V_{s_j-1}^{-1}}
    \;\le\; \frac{1}{\sqrt{N_{s_j-1}(X)}}
    \;\le\; \frac{1}{\sqrt{j-1}}.
    $$
    Combining with the threshold condition gives, for every $j\ge 2$,
    $$
    \frac{1}{\sqrt{j-1}} \;>\; \frac{C}{s_j^\beta}
    \quad\Longrightarrow\quad
    j-1 \;<\; \frac{s_j^{2\beta}}{C^2}
    \;\le\; \frac{t^{2\beta}}{C^2}.
    $$
    Thus $j \le 1 + t^{2\beta}/C^2$ for all such indices. Including the base case $j=1$ (which trivially satisfies the same bound), we obtain
    $$
    m_t(X) \;\le\; \frac{t^{2\beta}}{C^2} \,+\, 1 
    \;\le\; \frac{2\,t^{2\beta}}{C^2}
    \qquad\text{for all sufficiently large $t$.}
    $$
    Summing over $X\in\A$ then gives
    $$
    \sum_{X\in\A} m_T(X) \;\le\; \frac{2M}{C^2}\,T^{2\beta},
    $$
    which shows the total number of rounds that our algorithm performs the exploration. On all of the other rounds, we are playing based on an optimal regret minimization subroutine, like the CombLinUCB algorithm\cite{comblinucb-wen2015efficient}, where we can use their regret upper bound to bound our regret as 
    \begin{align} \label{eq: final regret upper bound}
        &Regret_T \big( \text{InfluenceCB}, \A, \theta^* \big)  \\
        &\leq  \tilde{\mathcal{O}} \big(c kd\sqrt{T}\big) + \frac{N}{C^2} T^{2 \beta} \in \tilde{\mathcal{O}} \left( T^ {2 \beta} \right),
    \end{align}
    where we use $\beta \geq \frac14$.

    We now prove that 
    \begin{align} \label{eq: normvinv upper bound}
        \forall X \in \A: \quad \|X\|_{V_T^{-1}} \le \frac{2C}{T^{\beta}}.
    \end{align}
    Since $V_t \succeq V_{t-1}$, the uncertainty $\|X\|_{V_t^{-1}}$ is monotonically decreasing in $t$.  
    If $\|X\|_{V_T^{-1}} > \tfrac{2C}{T^{\beta}}$, then for all $t < T$,
    $$
    \|X\|_{V_t^{-1}} > \frac{2C}{T^{\beta}} > \frac{C}{(T/2)^{\beta}},
    $$
    implying that for all $t \in [T/2, T]$, we have $u_t > C/t^{\beta}$ and the algorithm explores.  
    However, this would require
    $$
    \frac{T}{2} \le \frac{2M}{C^2}T^{2\beta},
    $$
    which fails for large $T$ for $\beta < \tfrac{1}{2}$, and for $\beta= \frac12$, the constant $C$ can be chosen to be large enough to fail this inequality.  
    Thus, for sufficiently large $T$, inequality~\eqref{eq: normvinv upper bound} must hold.  
    For small $T$, the constant in the final bound can be adjusted accordingly.
    
    From the self-normalized bound of~\cite{abbasi2011improved}, with probability at least $1 - \delta$, we have for all $X \in \A$:
    $$
    |X^\top(\hat\theta_T-\theta^*)|
    \;\le\;\beta_T(\delta)\,\sqrt{X^\top V_T^{-1} X},
    $$
    where $\beta_T(\delta)$ is defined in \eqref{eq: beta_T delta} and is logarithmic in the problem parameters.
    By setting $\delta = \frac{1}{kT}$ and taking the expectation, and combining this with~\eqref{eq: normvinv upper bound} implies
    $$
    |X^\top(\hat\theta_T-\theta^*)|
    \;\le\;\tilde{\mathcal{O}}\!\left(T^{-\beta}\right),
    $$
    and hence for any instance $(\A, \theta^*)$,
    \begin{align*}
    &RMSE_T\big(\text{InfluenceCB}, \A, \theta^*\big) \\
    &= \mathbb{E}\!\left[\sqrt{\frac{1}{M}\sum_{i=1}^{M}\bigl(X_i^\top(\hat\theta_T-\theta^*)\bigr)^2}\right]
    \;\in\;
    \tilde{\mathcal{O}} \!\left(T^{-\beta}\right).
    \end{align*}
    This completes the proof.
\end{proof}

\begin{figure}[t]
    \centering
     \begin{subfigure}
         \centering \includegraphics[width=\columnwidth]{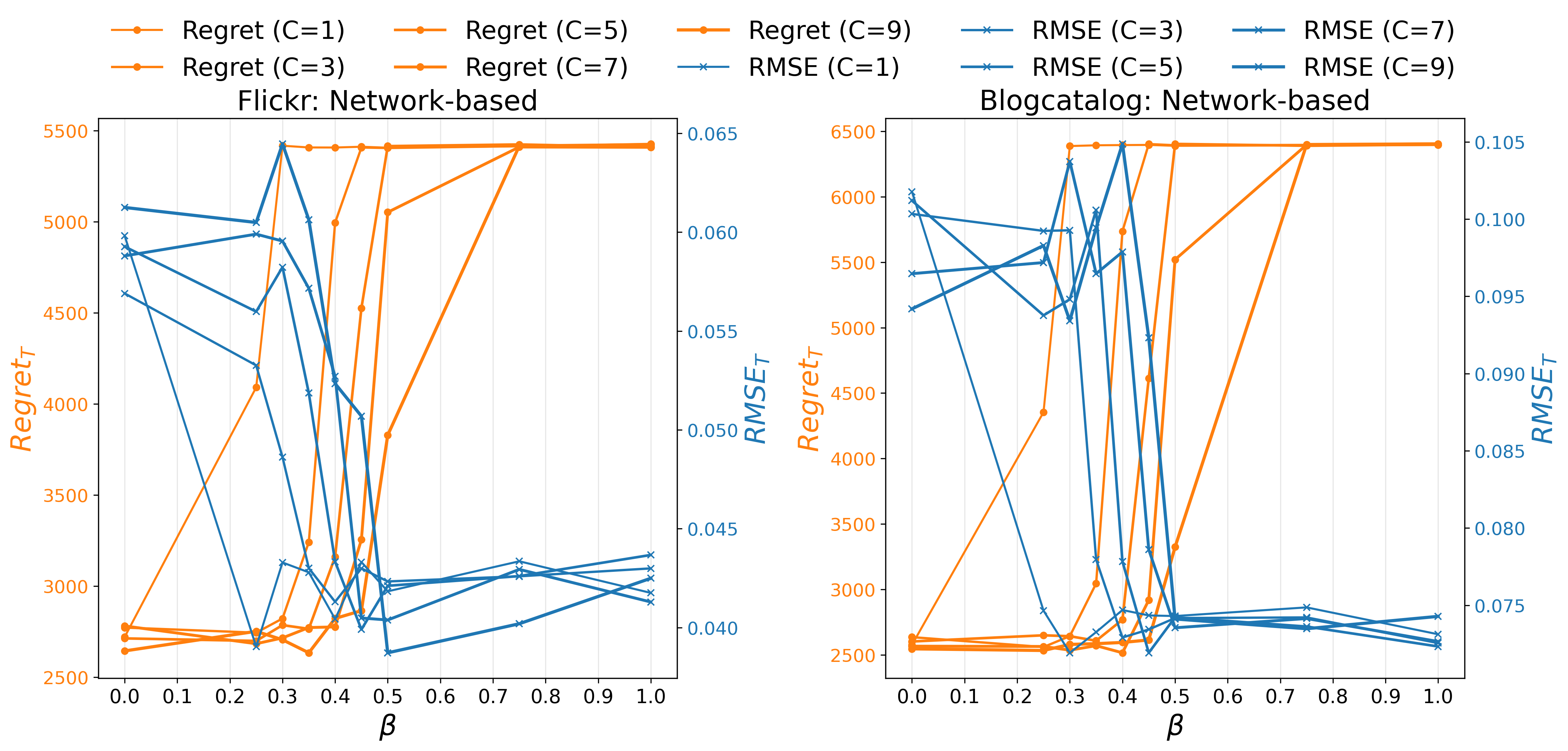}
     \end{subfigure}
     \vspace{-13 pt}
\caption{Trade-off between cumulative regret ($Regret_T$) and estimation error ($RMSE_T$) in $\Lhse(\beta, C)$ under a misspecified, non-linear influence function setting. Peer influence probabilities are generated via a feed-forward neural network to simulate model misspecification. Results on Flickr and BlogCatalog show that while $Regret_T$ trends remain consistent with the linear case, $RMSE_T$ exhibits more irregular patterns as $\beta$ and $C$ vary.}
     \label{fig-nonlinear_C_network}
\end{figure}

\subsection{Justification for a Linear Model for Influence Probabilities}
\label{section-justify-linear-model}
Peer influence is commonly modeled as an additive aggregation of social signals, such as weighted or cumulative exposure from neighbors~\cite{kempe-kdd-2003}. Linear models provide the canonical representation of such additive marginal contributions of features and are widely used in the influence and diffusion literature~\cite{he-neurips-2016}. Consistent with this perspective, statistical diffusion models typically introduce actor- and edge-level covariates through linear predictors, interpreting feature effects as additive on a latent influence scale, while nonlinearities arise from activation, saturation, or cascade dynamics rather than from individual feature interactions~\cite{myers-kdd-2012}. When features correspond to approximately independent drivers of persuasiveness or exposure—such as activity level, credibility, topical alignment, or tie-strength proxies—additivity constitutes a realistic first-order approximation of social response~\cite{aral-pnas-2009}. From a learning perspective, linear generalization enables identifiable peer effects and stable estimation in partially observed networks, and is particularly well suited to large-scale social graphs where the number of edges may reach millions or billions. Our choice is further aligned with IMLinUCB, which explicitly assumes linear generalization and is both statistically and computationally suitable for large-scale problems~\cite{vaswani-icml-2017}.

\subsection{Experiments with Non-linear Ground-truth Influence Model.}
\label{section-non-linear}
To capture non-linear interactions among edge features, we synthetically generate
peer influence probabilities using a feed-forward neural network. Given the feature
vector $e_{ij}.X \in \mathbb{R}^{256}$ for edge $e_{ij}$, the probability
$e_{ij}.p$ is computed as
\[
\begin{aligned}
h_1 = \tanh\!\left(W_1 e_{ij}.X + \mathbf{b}_1\right), \quad h_2 = \tanh\!\left(W_2h1 + \mathbf{b}_1\right),\\  \quad
z = W_3 h2 + b_3, \quad
e_{ij}.p = \sigma\!\left(\tfrac{z}{\tau}\right),
\end{aligned}
\]
where $W_1 \in \mathbb{R}^{128 \times 256}$, $W_2 \in \mathbb{R}^{64 \times 128}$,  and $W_3 \in \mathbb{R}^{1 \times 64}$
are weight matrices with entries drawn uniformly at random from $[-1,1]$,
$\mathbf{b}_1 \in \mathbb{R}^{128}$, $\mathbf{b}_2 \in \mathbb{R}^{64}$ and $b_2 \in \mathbb{R}$ are bias terms,
$\tanh(\cdot)$ denotes the hyperbolic tangent activation, and $\sigma(\cdot)$ is
the sigmoid function ensuring $e_{ij}.p \in (0,1)$. The scalar $\tau = 4.0$ is a
temperature parameter that controls the sharpness of the probability distribution. 

This experiment follows the same setup as Section~\ref{exp-1} but employs the synthetically generated peer influence probabilities described above to investigate how the generation model misspecification impacts the trade-off between cumulative regret \(Regret_T\) and estimation error \(RMSE_T\). The observed trade-off are presented in Figure~\ref{fig-nonlinear_C_network}. The results exhibit a similar overall pattern to those in Section~\ref{exp-1}:
$Regret_T$ behaves as expected for the most part but $RMSE_T$ has a more idiosyncratic behavior. A similar pattern is observed in the neighbor-based experiments.

\begin{table}[t]
\centering
\caption{$Regret_T$ and $RMSE_T$ achieved by $\Lhse(\beta, \mathcal{O})$ on the Flickr, BlogCatalog, and Hateful datasets with neighbor-based edge selection  (lower is better).}
\label{table-neighber-based-half}
\resizebox{\columnwidth}{!}{
\begin{tabular}{lcccccc}
\toprule
\multirow{2}{*}{\textbf{Method}} 
& \multicolumn{2}{c}{\textbf{Flickr}} 
& \multicolumn{2}{c}{\textbf{BlogCatalog}} 
& \multicolumn{2}{c}{\textbf{Hateful}} \\
\cmidrule(lr){2-3} \cmidrule(lr){4-5} \cmidrule(lr){6-7}
 & $Regret_T$  & $RMSE_T$  
 & $Regret_T$  & $RMSE_T$ 
 & $Regret_T$  & $RMSE_T$  \\
\midrule
\multicolumn{7}{c}{\textit{Static Models}} \\
\midrule
Random & 2400 & 0.3296 & 3009 & 0.3018 & 9610 & 0.2981 \\
Similarity & 2817 & 0.3439 & 3075 & 0.4757 & 9588 & 0.4993 \\
Link Prediction (Ridge) & 1338 & 0.1066 & 3215 & 0.1810 & 6679 & 0.1848 \\
Link Prediction (GraphSAGE) & 1407 & 0.4325 & 2950 & 0.2430 & 8792 & 0.4470 \\
Link Prediction (NCNC) & 1807 & 0.2507 & 2305 & 0.1539 & 6634 & 0.4120 \\
SOTA Recommender (ContextGNN)    & 1773 & 0.2370 & 2496 & 0.2125 & 6236 & 0.3550 \\
\midrule
\multicolumn{7}{c}{\textit{Bandit Models (CMAB)}} \\
\midrule
$LinUCB$ & 778 & 0.0542 & 654 & 0.0599 & 1057 & 0.0107 \\
$EENet$ & 655 & 0.0861 & 826 & 0.1001 & 5418 & 0.0664 \\
$GNB$ & 557 & 0.0785 & 785 & 0.0830 & 4267 & 0.0525\\
$\textit{Takemura et al.}$ & 413 & 0.0453 & 541 & 0.0555 & 646 & 0.0114\\
\midrule
\multicolumn{7}{c}{\textit{Online Learning}} \\
\midrule
$SGD_{Explore}$ & 2397 & 0.1366 & 3010 & 0.0779 & 9615 & 0.0500 \\
$SGD_{Exploit}$ & 424 & 0.1154 & 498 & 0.1574 & 2612 & 0.0536 \\
\midrule
\multicolumn{7}{c}{\textit{Ours: $\Lhse$ (InfluenceCB)}} \\
\midrule
$\Lhse(0.25,\text{ Regret})$ & 772 & 0.0486 & 596 & 0.0616 & 1049 & 0.0108 \\
$\Lhse(0.25,\text{ RMSE})$ & 936 & 0.0405 & 845 & 0.0483 & 1688 & 0.0105 \\
$\Lhse(0.50,\text{ Regret})$ & 1378 & 0.0332 & 1062 & 0.0427 & 5800 & 0.0102 \\
$\Lhse(0.50,\text{ RMSE})$ & 2289 & 0.0365 & 2495 & 0.0384 & 10095 & 0.0101 \\
\bottomrule
\end{tabular}
}
\end{table}


\begin{table}[t]
\centering
\caption{$Regret_T$ and $RMSE_T$ achieved on the \textbf{BlogCatalog} dataset with network-based edge selection.}
\vspace{-8pt}
\label{table-blogcatalog-ablation-lints}
\resizebox{\columnwidth}{!}{
\begin{tabular}{lcccc}
\toprule
\textbf{Method} 
& $Regret_T$ $\downarrow$ 
& $RMSE_T$ $\downarrow$ 
& $ECE_T$ $\downarrow$ 
& $NDCG@5$ $\uparrow$ \\
\midrule
\multicolumn{5}{c}{\textit{Bandit Baseline}} \\
\midrule
$LinUCB$ 
& 1275 & 0.0993 & 0.0865 & 0.9168 \\
\midrule
\multicolumn{5}{c}{\textit{Ours: InfluenceCB ($\Lhse$)}} \\
\midrule
$\Lhse (0.25, Regret)$ 
& \textbf{1141} & \textbf{0.1038} & \textbf{0.0880} & \textbf{0.9251} \\

$\Lhse (0.25, RMSE)$ 
& \textbf{1337} & \textbf{0.0644} & \textbf{0.0492} & \textbf{0.9130} \\

$\Lhse (0.50, Regret)$ 
& \textbf{2198} & \textbf{0.0453} & \textbf{0.0269} & \textbf{0.8597} \\

$\Lhse (0.50, RMSE)$ 
& \textbf{3780} & \textbf{0.0334} & \textbf{0.0149} & \textbf{0.7597} \\
\bottomrule
\end{tabular}
}
\end{table}

\subsection{Ablation: LinTS-Based Exploitation}
\label{ablation-LinTS}
To assess whether the empirical behavior of $\Lhse$ depends on the specific choice of linear CMAB exploitation, we conduct an ablation in which only the reward-guided exploitation component is replaced with Linear Thompson Sampling (LinTS), while all other components of the algorithm remain unchanged in the network-based setting. In particular, the uncertainty-guided exploration mechanism, candidate action pools, update rules, and the trade-off parameter $\beta$ are kept fixed. During exploitation rounds, we sample a parameter vector from the Gaussian posterior approximation induced by the linear model and rank candidate edges using the sampled predicted influence scores, selecting the top-$k$ edges accordingly. This replaces the LinUCB-based ranking used in the original exploitation step. The exploration rounds continue to prioritize uncertainty reduction as in the base algorithm. This ablation isolates the effect of the exploitation strategy and allows us to evaluate whether the observed $Regret$--$RMSE$ trade-offs persist under LinTS-based exploitation. The results in Table~\ref{table-blogcatalog-ablation-lints} show qualitatively similar trade-offs to those obtained with LinUCB-based exploitation, suggesting that the empirical findings of $\Lhse$ are robust to the choice of linear exploitation mechanism.


\subsection{Formal Definitions of Evaluation Metrics}
\label{section-detailed-metrics}
We evaluate the performance of the $\Lhse$ using the following metrics:
{\subsubsection{$Regret_T$.} 
To evaluate regret, we compute the expected activations of the \(k\) recipient nodes influenced by their corresponding source nodes across the \(k\) \(\{e_{ij}\}\), selected by \(\Lhse\) in round \(t\). These activations are estimated through simulations using the ground-truth peer influence probabilities. We then compare this value against the expected activations obtained from the top-\(k\) edges \(\{e^*_{ij}\}\), representing the maximum achievable expected activations under the same simulation setup in round \(t\). The cumulative regret over \(T\) rounds, \(Regret_T\), is then computed following the formulation defined in Section~\ref{CMAB-setup}.
}
\subsubsection{$RMSE_T$} 
To evaluate the accuracy of the learned peer influence probabilities, we compute the root mean squared error in round $T$, $RMSE_T$, over a randomly sampled subset of edges \(\overline{E} \subset E\) from the network. The estimation error is calculated as: $RMSE_T = \sqrt{\frac{\sum_{e_{ij} \in \overline{E}}{(\hat{e}_{ij}.p - e_{ij}.p)^2}}{|\overline{E}|}}$, where $\hat{e}_{ij}.p$ = $min(1, max(0, {X}^{\top}\theta^*))$ is the estimated peer influence probability in round $T$ and , computed following the formulation defined in Section~\ref{CMAB-setup}.

\begin{figure*}[t]
\centering
\includegraphics[width=\textwidth]{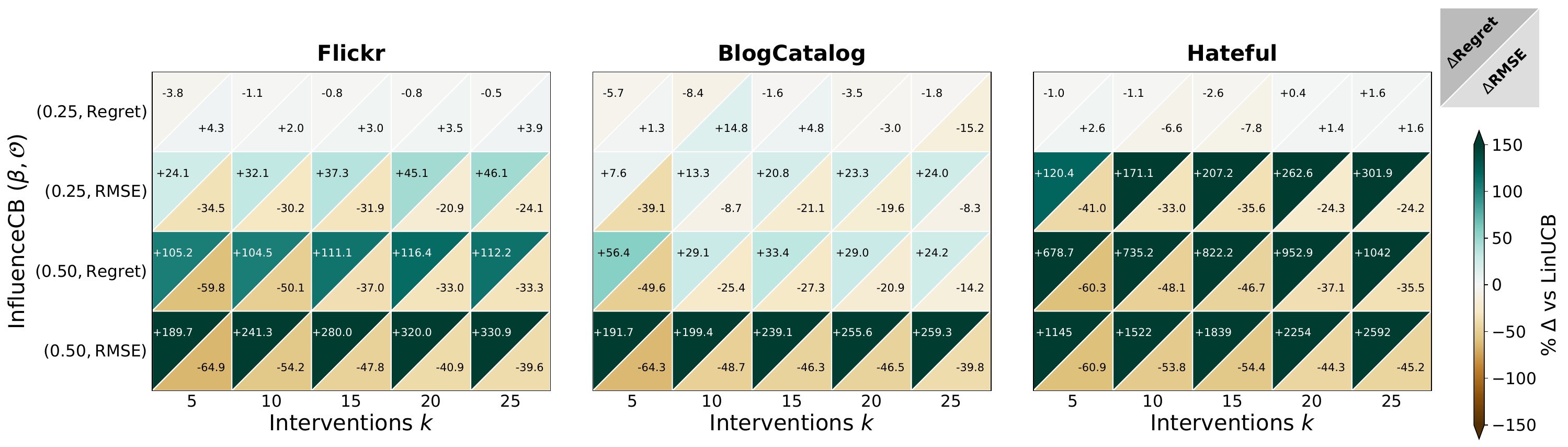}
\vspace{-13 pt}
\caption{Effect of the number of interventions $k$ on the
$\mathit{Regret}_T$--$\mathit{RMSE}_T$ trade-off across the Flickr,
BlogCatalog, and Hateful datasets (network-based edge selection). Each
cell reports the percentage change (\%) of an
$\mathit{InfluenceCB}(\beta, \mathcal{O})$ variant relative to
$\mathit{LinUCB}$. Cells are split diagonally: the upper-left triangle
shows $\Delta\mathit{Regret}_T$ and the lower-right triangle shows
$\Delta\mathit{RMSE}_T$. Color encodes magnitude on a shared diverging
scale clipped at $\pm350\%$ (cells beyond this range saturate and are
marked with arrows on the colorbar; exact values are printed in every
cell). A negative $\Delta\mathit{RMSE}_T$ indicates an estimation
improvement and a positive $\Delta\mathit{Regret}_T$ a regret cost. The
$k=5$ slice matches the network-based results in
Table~\ref{table-network-based-half}. When RMSE is prioritized
($\beta=0.5$), the estimation-error reduction shrinks as $k$ grows while
the regret cost rises sharply.}
\label{fig:vary-k-heatmap}
\end{figure*}

\subsubsection{Expected Calibration Error (ECE).}
ECE measures how well the predicted peer influence probabilities are calibrated against the ground-truth influence probabilities. After $T$ rounds, we collect predicted probabilities $\hat{p}_{ij}$ for all held-out edges $e_{ij} \in \overline{E}$ and partition them into $B$ equal-width bins $\{\mathcal{B}_b\}_{b=1}^{B}$ over $[0,1]$. For each bin $\mathcal{B}_b$, let
\[
\text{conf}(\mathcal{B}_b) = \frac{1}{|\mathcal{B}_b|} \sum_{e_{ij} \in \mathcal{B}_b} \hat{p}_{ij},
\qquad
\text{acc}(\mathcal{B}_b) = \frac{1}{|\mathcal{B}_b|} \sum_{e_{ij} \in \mathcal{B}_b} p_{ij}.
\]
The Expected Calibration Error is defined as
\[
\mathrm{ECE}
=
\sum_{b=1}^{B}
\frac{|\mathcal{B}_b|}{|\overline{E}|}
\left|
\text{acc}(\mathcal{B}_b) - \text{conf}(\mathcal{B}_b)
\right|.
\]
Lower $\mathrm{ECE}$ indicates better calibration of the learned influence probabilities.

\subsubsection{Normalized Discounted Cumulative Gain (NDCG@k).}
NDCG@k evaluates the quality of the top-$k$ edges selected by the algorithm. We treat the ground-truth influence probabilities $p_{ij}$ as graded relevance scores. Given a predicted ranking of $k$ edges, the discounted cumulative gain is
\[
\mathrm{DCG@}k
=
\sum_{r=1}^{k}
\frac{2^{\,p_{(r)}} - 1}{\log_2(r+1)},
\]
where $p_{(r)}$ denotes the ground-truth influence probability of the edge ranked at position $r$. The normalized score is obtained by dividing by the DCG of the ideal ranking induced by the top-$k$ edges with the largest $p_{ij}$:
\[
\mathrm{NDCG@}k
=
\frac{\mathrm{DCG@}k}{\mathrm{IDCG@}k}.
\]
Higher NDCG@k indicates better alignment between the algorithm’s selected edges and the true influence ordering. We report NDCG@k averaged across rounds.

\subsection{Effect of k on the Regret-RMSE tradeoff}
Figure~\ref{fig:vary-k-heatmap} shows the improvement of $\mathit{InfluenceCB}(\beta, \mathcal{O})$ over $LinUCB$ in RMSE and Regret for different values of $k$, $\beta$, and optimization objectives $\mathcal{O}$. 

\subsection{Effect of feature dimension $d$ on the trade-off in $\mathit{InfluenceCB}(\beta,\mathcal{O})$.}
\label{sec:effect-of-d}
\begin{figure*}[t]
\centering
\includegraphics[width=\textwidth]{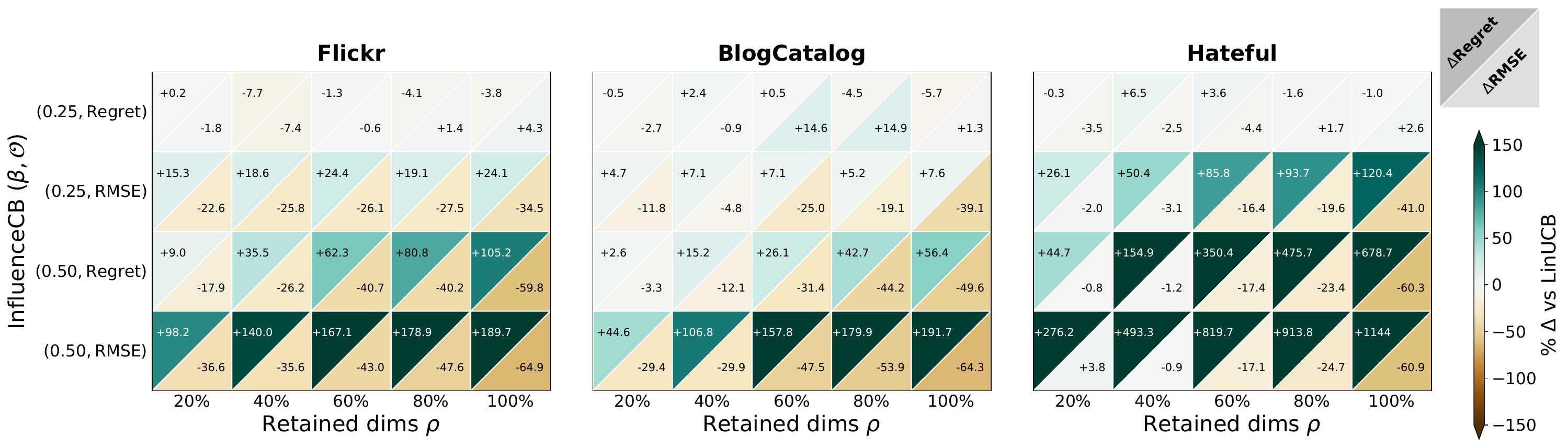}
\vspace{-13 pt}
\caption{Effect of feature dimension on the $\mathit{Regret}_T$--$\mathit{RMSE}_T$
trade-off across datasets. Each cell reports the percentage change (\%) of an
$\mathit{InfluenceCB}(\beta, \mathcal{O})$ variant relative to $\mathit{LinUCB}$
(network-based, $k=5$) at a given fraction $\rho$ of retained feature dimensions.
Cells are split diagonally: the upper-left triangle shows $\Delta\mathit{Regret}_T$
and the lower-right triangle shows $\Delta\mathit{RMSE}_T$. Color encodes magnitude
on a shared diverging scale; a negative $\Delta\mathit{RMSE}_T$ (brown) indicates an
estimation improvement, while a positive $\Delta\mathit{Regret}_T$ (teal) indicates a
regret cost. Across both datasets, the regret--estimation trade-off becomes more
pronounced as $\rho$ increases.}
\label{fig:vary-d-heatmap}
\end{figure*}

We randomly retain a fraction $\rho \in \{0.2, 0.4, 0.6, 0.8, 1.0\}$ of the 256
feature dimensions and report the percentage change in $\mathit{Regret}_T$ and
$\mathit{RMSE}_T$ of each $\mathit{InfluenceCB}$ variant relative to $\mathit{LinUCB}$
across the three datasets (network-based, $k=5$); a negative $\Delta\mathit{RMSE}_T$
indicates an estimation improvement and a positive $\Delta\mathit{Regret}_T$ a regret
cost (Figure~\ref{fig:vary-d-heatmap}). The same qualitative pattern holds on all three
datasets: as $\rho$ grows from $20\%$ to $100\%$, the $\mathit{RMSE}_T$ reduction widens
and the corresponding regret cost grows, with the effect scaling in $\beta$.
$\mathit{InfluenceCB}(0.25, \mathrm{Regret})$ stays close to $\mathit{LinUCB}$ on both
metrics (within $\pm 8\%$ on Flickr and Hateful; on BlogCatalog its regret remains flat
while $\Delta\mathit{RMSE}_T$ drifts up to about $+15\%$). The scaling is sharpest for the
high-$\beta$ variants: on Flickr, $\mathit{InfluenceCB}(0.50, \mathrm{Regret})$ moves from
$-17.88\%$ to $-59.81\%$ $\mathit{RMSE}_T$ while its regret cost rises from $+8.97\%$ to
$+105.22\%$, and $\mathit{InfluenceCB}(0.50, \mathrm{RMSE})$ from $-36.57\%$ to $-64.87\%$
$\mathit{RMSE}_T$ at $+98.15\%$ to $+189.65\%$ regret. $\mathit{InfluenceCB}(0.25, \mathrm{RMSE})$
offers the most stable operating point, trading roughly $+15\%$ to $+24\%$ regret for
$-22\%$ to $-34\%$ $\mathit{RMSE}_T$ across all $\rho$ on Flickr. The trend is most
pronounced on the large, sparse Hateful graph, where the regret cost of
$\mathit{InfluenceCB}(0.50, \mathrm{Regret})$ escalates from $+44.70\%$ to $+678.74\%$ as
$\rho$ increases, for an $\mathit{RMSE}_T$ gain widening from $-0.76\%$ to $-60.26\%$.
Overall, the regret--estimation trade-off becomes more pronounced as $d$ increases, and
the cost side grows faster than the estimation return at higher dimensions.

\begin{table*}[t]
\centering
\caption{$Regret_T$, $RMSE_T$, $ECE_T$, and $NDCG@5$ achieved by $\Lhse(\beta, \mathcal{O})$ 
on the Flickr, BlogCatalog, and Hateful datasets with network-based edge selection.}
\vspace{-10pt}
\label{table-network-based-full}
\resizebox{\textwidth}{!}{
\begin{tabular}{lcccccccccccc}
\toprule
\multirow{2}{*}{\textbf{Method}} & \multicolumn{4}{c}{\textbf{Flickr}} & \multicolumn{4}{c}{\textbf{BlogCatalog}} & \multicolumn{4}{c}{\textbf{Hateful}}\\
\cmidrule(lr){2-5} \cmidrule(lr){6-9} \cmidrule(lr){10-13}
 & $Regret_T$ $\downarrow$ & $RMSE_T$ $\downarrow$ & $ECE_T$ $\downarrow$ & $NDCG@5$ $\uparrow$ & $Regret_T$ $\downarrow$ & $RMSE_T$ $\downarrow$ & $ECE_T$ $\downarrow$ & $NDCG@5$ $\uparrow$ & $Regret_T$ $\downarrow$ & $RMSE_T$ $\downarrow$ & $ECE_T$ $\downarrow$ & $NDCG@5$ $\uparrow$ \\
\midrule
\multicolumn{13}{c}{\textit{Static Models}} \\
\midrule
Random        & 5428 & 0.3296 & 0.2719 & 0.7688 & 5419 & 0.3018 & 0.2490 & 0.6528 & 73226 & 0.2981 & 0.2518 & 0.7263 \\
Similarity    & 5506 & 0.3440 & 0.3398 & 0.7645 & 5590 & 0.4757 & 0.4682 & 0.6413 & 82962 & 0.4993 & 0.4946 & 0.6922 \\
Link Prediction (Ridge)         & 5090 & 0.1066 & 0.0629 & 0.7847 & 6426 & 0.1810 & 0.1297 & 0.5955 & 42384 & 0.1848 & 0.1658 & 0.8356 \\
Link Prediction (GraphSAGE)     & 4772 & 0.4325 & 0.4254 & 0.7973 & 5616 & 0.2430 & 0.2056 & 0.6492 & 41568 & 0.4470 & 0.4425 & 0.8323 \\
Link Prediction (NCNC)     & 4579 & 0.2325 & 0.2629 & 0.8074 & 4918 & 0.2130 & 0.1822 & 0.6621 & 39850 & 0.4120 & 0.4015 & 0.8420 \\
SOTA Recommender (ContextGNN)     & 4650 & 0.2550 & 0.2420 & 0.8220  & 4800 & 0.2250 & 0.1880 & 0.6510 & 40500 & 0.3950 & 0.3720 & 0.8550 \\
\midrule
\multicolumn{13}{c}{\textit{Bandit Models (CMAB)}} \\
\midrule
$LinUCB$        & 2165 & 0.0632 & 0.0532 & 0.9049 & 1275 & 0.0993 & 0.0865 & 0.9168 & 5979 & 0.0156 & 0.0110 & 0.9756 \\
$EENet$         & 1884 & 0.1164 & 0.0761 & 0.9225 & 1683 & 0.1302 & 0.0821 & 0.9330 & 9635 & 0.0658 & 0.0466 & 0.9613 \\
$GNB$ 
& 1314 & 0.0719 & 0.0560 & 0.9463
& 1586 & 0.0945 & 0.0760 & 0.9285
& 8669 & 0.0417 & 0.0290 & 0.9645 \\

$\textit{Takemura et al.}$ 
& 952 & 0.0681 & 0.0540 & 0.9615
& 912 & 0.0990 & 0.0800 & 0.9370
& 2829 & 0.0267 & 0.0185 & 0.9798 \\

\midrule
\multicolumn{13}{c}{\textit{Online Learning}} \\
\midrule
$SGD_{Explore}$ & 5425 & 0.1232 & 0.1168 & 0.7688 & 5432 & 0.0773 & 0.0293 & 0.6523 & 73297 & 0.0433 & 0.0285 & 0.7261 \\
$SGD_{Exploit}$  & 822 & 0.1006 & 0.0866 & 0.9668 & 1015 & 0.1007 & 0.0814 & 0.9340 & 9441 & 0.0924 & 0.0787 & 0.9634 \\
\midrule
\multicolumn{13}{c}{\textit{Ours: $\Lhse$}} \\
\midrule
$\Lhse (0.25, Regret)$   & 2083 & 0.0659 & 0.0526 & 0.8821 & 1202 & 0.1006 & 0.0794 & 0.9012 & 5917 & 0.0160 & 0.0118 & 0.9762 \\
$\Lhse (0.25, RMSE)$ 
& 2686 & 0.0414 & 0.0321 & 0.8843 & 1372 & 0.0605 & 0.0458 & 0.9109 & 13179 & 0.0092 & 0.0042 & 0.9504\\
$\Lhse (0.50, Regret)$   & 4443 & 0.0254 & 0.0104 & 0.8154 & 1994 & 0.0500 & 0.0340 & 0.8721 & 46561 & 0.0062 & 0.0013 & 0.8292 \\
$\Lhse (0.50, RMSE)$   & 6271 & 0.0222 & 0.0055 & 0.7396 & 3719 & 0.0354 & 0.0160 & 0.7629 & 74399 & 0.0061 & 0.0014 & 0.7250 \\
\bottomrule
\end{tabular}
}
\end{table*}
\end{document}